\documentclass[runningheads]{llncs}

\usepackage[table,xcdraw]{xcolor}
\usepackage{graphicx}
\usepackage{booktabs}
\usepackage{pifont}
\usepackage{eccv}
\usepackage[accsupp]{axessibility}  

\usepackage[pagebackref,breaklinks,colorlinks,citecolor=eccvblue]{hyperref}

\usepackage{orcidlink}
\usepackage{multirow}
\usepackage{tcolorbox}
\usepackage{amssymb}

\newcommand{\ourmodel}[0]{\text{TheaterGen}}
\newcommand{\ourdataset}[0]{\text{CMIGBench}}
\begin{document}

\title{TheaterGen \includegraphics[width=0.6cm]{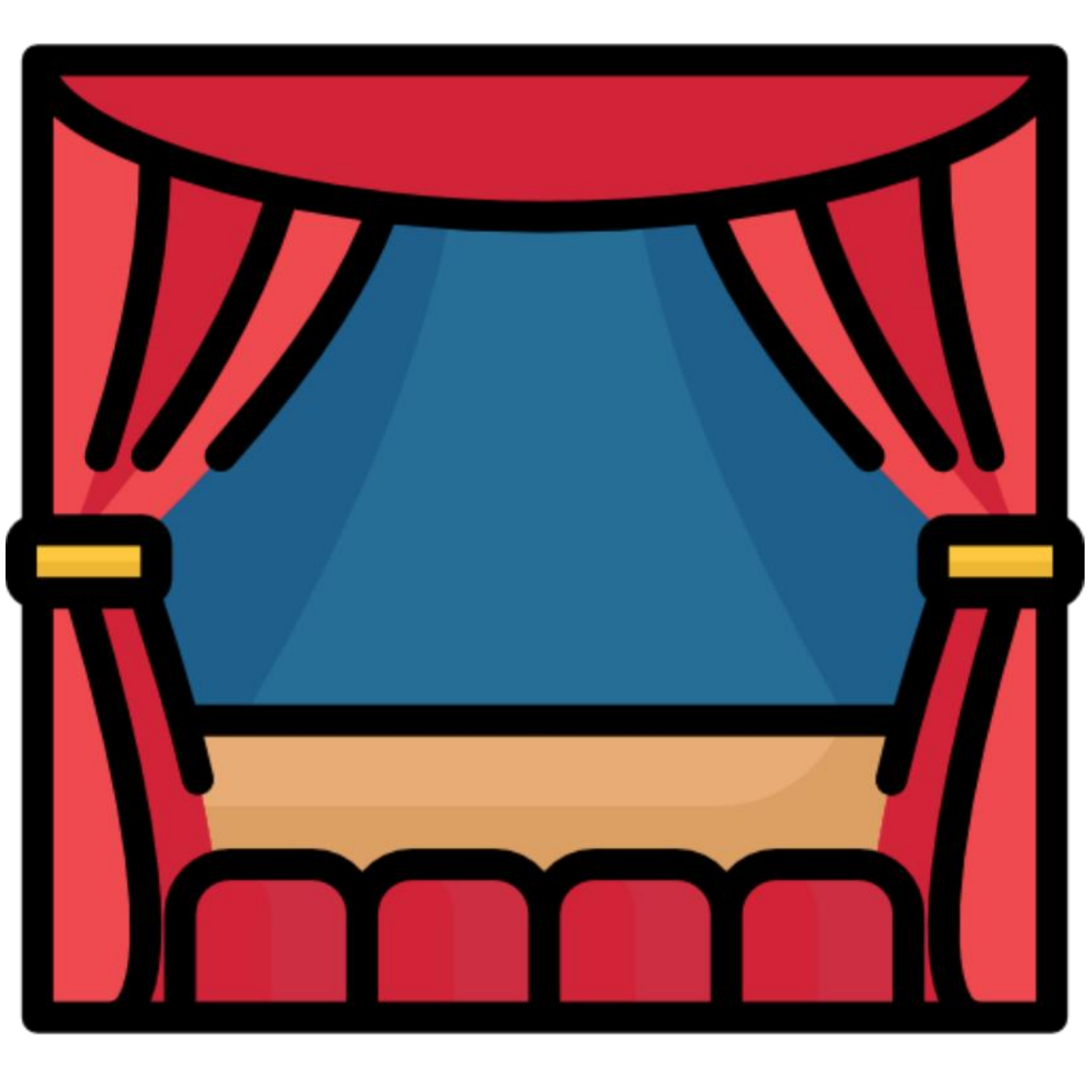}: Character Management with \\ LLM for Consistent Multi-turn Image Generation} 

\titlerunning{Theatergen}

\author{Junhao Cheng\inst{1} \and Baiqiao Yin\inst{1} \and Kaixin Cai\inst{1} \and Minbin Huang\inst{2} \and Hanhui Li\inst{1} \and Yuxin He\inst{1} \and Xi Lu\inst{1} \and Yue Li\inst{1} \and Yifei Li\inst{1} \and Yuhao Cheng\inst{3} \and Yiqiang Yan\inst{3} \and Xiaodan Liang\inst{1,*}}

\authorrunning{Cheng et al.}

\institute{Shenzhen Campus of Sun Yat-sen University \and
The Chinese University of Hong Kong\\
\and Lenovo Research \\ * Corresponding Author, \{xdliang328\}@gmail.com \\
\url{https://howe140.github.io/theatergen.io/}}

\maketitle
\vspace{-1em}
\begin{abstract}
Recent advances in diffusion models can generate high-quality and stunning images from text. However, multi-turn image generation, which is of high demand in real-world scenarios, still faces challenges in maintaining semantic consistency between images and texts, as well as contextual consistency of the same subject across multiple interactive turns. To address this issue, we introduce \textbf{TheaterGen}, a training-free framework that integrates large language models (LLMs) and text-to-image (T2I) models to provide the capability of multi-turn image generation. Within this framework, LLMs, acting as a "\textbf{Screenwriter}", engage in multi-turn interaction, generating and managing a standardized prompt book that encompasses prompts and layout designs for each character in the target image. Based on character prompts and layouts, we generate a list of character images and extract guidance information from them, akin to the "\textbf{Rehearsal}". Subsequently, through incorporating the prompt book and guidance information into the reverse denoising process of T2I diffusion models, we generate the final image, as conducting the "\textbf{Final Performance}". With the effective management of prompt books and character images, TheaterGen significantly improves semantic and contextual consistency in synthesized images. Furthermore, we introduce a dedicated benchmark, CMIGBench (Consistent Multi-turn Image Generation Benchmark) with 8000 multi-turn instructions. Different from previous multi-turn benchmarks, CMIGBench does not define characters in advance, and hence it is of great diversity. Both the tasks of story generation and multi-turn editing are included on CMIGBench for comprehensive evaluation. Extensive experimental results show that TheaterGen outperforms state-of-the-art methods significantly. For example, it raises the performance bar of the cutting-edge Mini DALL·E 3 model by 21\% in average character-character similarity and 19\% in average text-image similarity. Our code and CMIGBench can be found in supplementary materials. Our codes and benchmark will be available at \url{https://github.com/donahowe/Theatergen}.

\keywords{Diﬀusion models \and Consistency \and Multi-turn image generation}
\end{abstract}

\section{Introduction}

\begin{figure*}[!tb]
  \centering
  \includegraphics[width=0.95\textwidth]{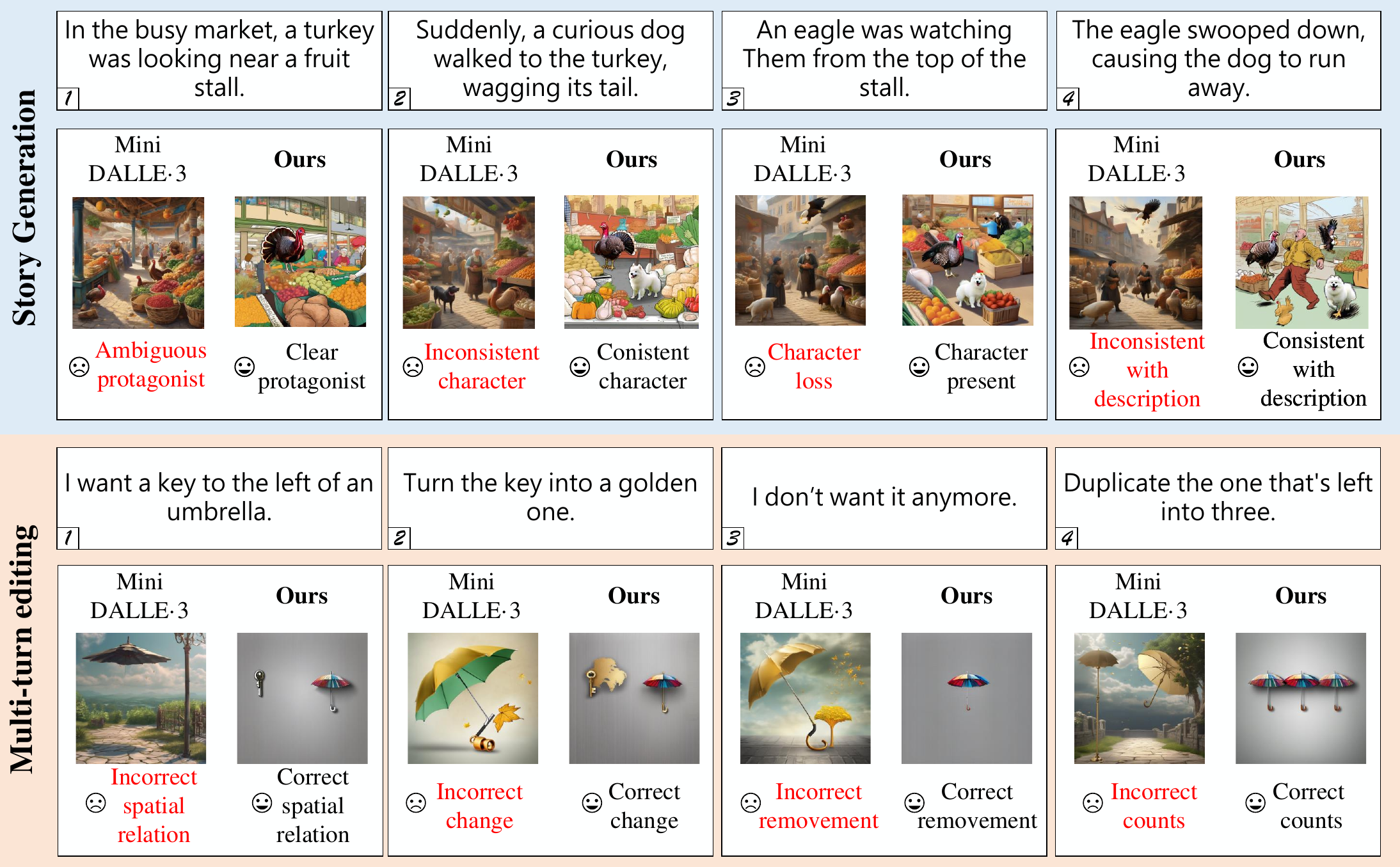}
  \caption{Visual comparison between Mini DALL·E 3 \cite{Mini-DALLE3} and our proposed TheaterGen in multi-turn story generation and multi-turn editing.}
  \label{fig:intro}
    \vspace{-2em}
\end{figure*}

In recent years, significant progress has been observed in text-to-image (T2I) generation tasks, attributed to the advancements in diffusion models \cite{sd}. State-of-the-art T2I models \cite{gafni2022make,lian2023llm} are capable of synthesizing diverse and detailed images that are truly astonishing. However, there is a growing need for multi-turn image generation. As shown in Figure \ref{fig:intro}, this encompasses the demand to produce coherent narrative images, such as sequential storylines, across multi-turn dialogue. Additionally, there may exist a requirement for iterative image refinement, necessitating multi-turn editing feedback to achieve satisfactory results. These demands present substantial challenges to text comprehension capability and the maintenance of consistency in multi-turn image generation within T2I models.

To address this challenge, existing work attempts to utilize large language models (LLM) \cite{Mini-DALLE3,} to extract pertinent information from multi-turn dialogues, thereby guiding diffusion models. Mini DALLE·3 \cite{Mini-DALLE3} references to the recent T2I model DALLE 3 \cite{dalle3}, and employs LLM to generate image descriptions or image embeddings during multi-turn dialogues, subsequently supplying these to the T2I model for image generation. MiniGPT-5 \cite{MiniGPT5} equips LLMs with the current story texts and corresponding images to generate the subsequent image embedding of the story. SEED-LLaMA \cite{SEED} integrates LLM with the SEED image tokenizer, enabling it to process and generate text and images interchangeably, leading to enhanced multimodal comprehension and generation capabilities.


Despite these advancements in handling complex texts within multi-turn dialogues, as shown in Figure \ref{fig:intro}, they still grapple with two primary issues: (1) \textbf{Semantic Consistency} - existing methods encounter difficulties in processing complex descriptions such as spatial relations, quantities, or referential expressions such as "it, they" during multi-turn dialogues, resulting in the generation of images that are semantically inconsistent with user requests. (2) \textbf{Contextual  Consistency} - during image generation within multi-turn dialogues, the same entity often struggles to maintain consistent features across different turns, and may even be forgotten. For example, the same dog looks different in different turns without user edits.


To tackle the above issues, we propose TheaterGen in this paper. In the context of multi-turn image generation, the consideration of information from preceding turns is important, which is an ability that current T2I models lack. Inspired by \cite{LMD}, we develop an LLM-based character designer that leverages an LLM to establish and manage a standardized prompt book comprising prompts and bounding boxes for each foreground character. This method serves to enhance comprehension and arrangement of user intents within a multi-turn interactive environment, effectively disentangling the intricate generation task into amalgamations of multiple characters. This process results in the production of more manageable text for multi-turn generation tasks, thereby promoting the achievement of semantic consistency.

To effectively leverage the prompt book, TheaterGen generates and maintains a reference image for each character. To ensure contextual consistency, upon the introduction of a new character, we generate an image using the character description with an off-the-shelf T2I model and preserve it as the reference of the character. The reference image is considered as the contextual prior, and will be amalgamated with the character prompt of the ongoing turn to derive the corresponding on-stage character image. With on-stage character images of multiple characters, TheaterGen further extracts lineart and latent guidance along with the bounding boxes from the prompt book to generate the final image with high consistency. In this way, even without extra training on a specific dataset, TheaterGen can align existing LLMs with T2I models to achieve consistent multi-turn image generation.

To evaluate both the semantic and contextual consistency in multi-turn image generation, we propose a new benchmark called CMIGBench. Our benchmark focuses on two common types found in multi-turn image generation: story generation and multi-turn editing. It consists of 8000 multi-turn scripted dialogues, with each dialogue containing 4 turns of natural language instructions. As in real-world scenarios, generation models might not know specific characters before the conversation starts, CMIGBench does not pre-define the characters. This makes our benchmark differ from previous benchmarks significantly and avoids the risk of loss in diversity. We compare \ourmodel ~with several cutting-edge models on CMIGBench, and the qualitative and quantitative results show that our method outperforms existing methods by large margins. 


To summarize, our main contributions are as follows:
\begin{enumerate}
    \item We propose TheaterGen, which is a training-free framework that utilizes a large language model to drive a text-to-image generation model, effectively addressing the issues of semantic consistency and contextual consistency in multi-turn image generation tasks without specialized training.
    \item TheaterGen can engage in multi-turn natural language interactions with users to accomplish tasks such as story generation and multi-turn editing.
    \item We propose a new benchmark, CMIGBench, to evaluate both the semantic and contextual consistency in multi-turn image generation and demonstrate the superior performance of TheaterGen.
\end{enumerate}

\section{Related Works}

\subsection{Text to Image Generation} 

Text-to-image generation is currently a widely researched and applied field. Mainstream image generation methods include Variational AutoEncoder \cite{tibebu2022text}, flow-based models \cite{lu2020structured}, Generative Adversarial Networks (GANs) \cite{GAN, GANsu, styleGAN, styleGAN2}, and diffusion models \cite{ddpm, sd, sdxl}. Recently, the stunning results of generating images from text descriptions using diffusion models have led to the emergence of many excellent diffusion-based image generation models \cite{ipadapter, controlnet, DALL·E2, DALL·E3}. These models encode text descriptions through multimodal models such as CLIP \cite{Clip, Align, ALBEF, Blip} and incorporate them into the image generation process through cross-modal attention mechanisms. Our TheaterGen primarily explores the potential for multi-turn image generation from text-to-image models.

\subsection{Multi-turn Image Generation with LLMs}

Text-to-image (T2I) generation models traditionally require precise and highly specific textual prompts, posing a challenge for natural language compatibility. Recent research efforts have focused on bridging this gap by integrating large language models (LLMs) with T2I generation models to enhance interaction through natural language to achieve multi-turn image generation\cite{SEED, MiniGPT5, DREAMLLM, LMD, AutoStory, TaleCrafter, GILL, huang2024dialoggen,cheng2024autostudio,cheng2025animegamer,luo2025object,cheng2024autostudio}. 
Mini DALL·E 3 extends the multi-turn interactive capabilities of its predecessors by categorizing user interaction into generation and editing modes. Diffusion GPT \cite{DiffusionGPT} introduces a "Tree-of-Thought of Models "\cite{ToM}  for model selection, guiding LLMs to parse user inputs and match the most suitable T2I model. Next GPT \cite{NextGPT} achieves multimodal alignment by encoding inputs across modalities uniformly, adjusting and decoding through LLMs. MiniGPT-5 \cite{MiniGPT5} uses LLM to understand images and text, inferring the next steps in the story and generating visual features to guide the generation of the next story image. In our work, TheaterGen uses LLM to maintain the layout and foreground character prompts in multi-turn interactions, thus achieving high-quality generation.

\subsection{Compositional Image Generation Task for Multi-turn Image Editing}

There exists another type of compositional image generation task \cite{DCG, GeNeVA, SSCR, LGAN} aimed at achieving more consistent completion of image editing tasks based on a given series of instructions without LLM. Among these efforts,  \cite{DCG} involves fine-tuning the editing instructions by incorporating attention mechanisms and introducing loss guidance. 
GeNeVA-GAN \cite{GeNeVA} uses a GRU \cite{GRU} to combine the previous condition and the current text embedding to produce the current context, guiding the generator to generate the current image from the previous round's image embedding. 
Finally, adversarial training with images from two rounds completes the construction of the iterative editor. 
SSCR \cite{SSCR} extends the GeNeVA-GAN \cite{GeNeVA} iterative editor by adding a GRU-based iterative interpreter, which generates semantic differences in text across different rounds as guidance. LatteGAN \cite{LGAN} utilizes attention for text guidance, introducing a visually guided language attention module  
to ensure images better align with editing instructions.

\section{Method}

\begin{figure}[tb]
  \centering
  \includegraphics[width=\textwidth]{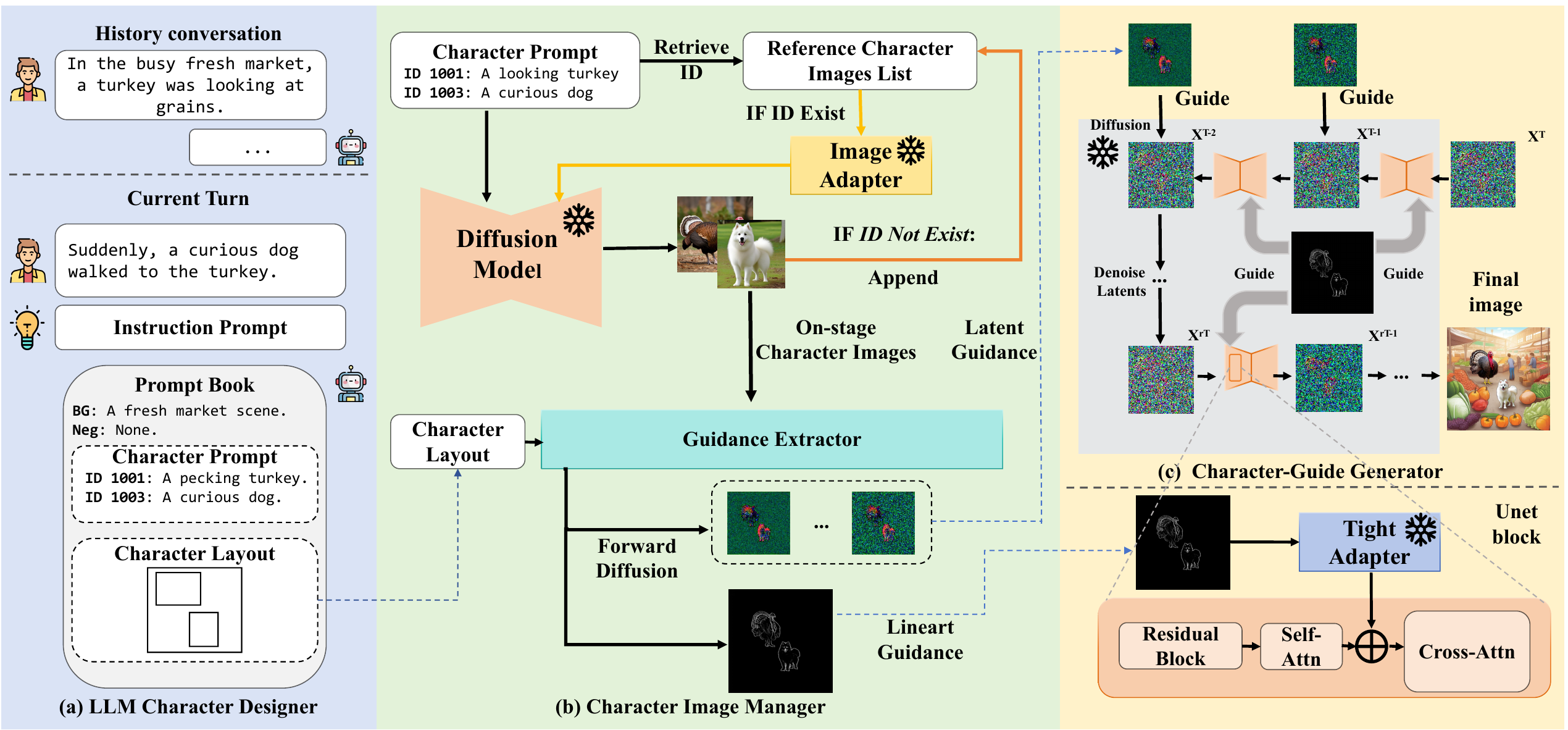}
  \caption{The overall structure of TheaterGen. TheaterGen utilizes three key components to generate an image in each interaction turn: (a) an LLM-based character designer that interacts with the user and maintains a structured prompt book for all character prompts and layouts, which serves as the ``\textbf{screenwriter}''; (b) a character image manager for ``\textbf{rehearsal}'', which generates character images and extracts guidance based on the prompt book; (c) a character-guided generator that conducts the ``\textbf{final performance}'', i.e., generates the final image for the current turn by combining the prompt book and guidance information.}
  \label{fig:model}
  \vspace{-1em}
\end{figure}

Figure \ref{fig:model} shows the overall framework of our proposed TheaterGen for multi-turn interactive image generation. We first introduce an LLM-based character designer (Section \ref{sec: LLM Layout generation}) to interact with the user and generate a prompt book consisting of character prompts and layouts. After that, a character image manager (Section \ref{sec:Local image generation}) is employed to generate on-stage character images and their corresponding lineart and latent guidance. At last, these two types of guidance and the prompt book are fed into a character-guided generator 
to synthesize the final image (Section \ref{sec:Global image generation}). In this way, our training-free framework can leverage the advantages of existing LLMs and T2I models in natural interaction and diverse image generation, while eliminating the computationally expensive training process. 

\subsection{LLM-based Character Designer}
\label{sec: LLM Layout generation}
Although existing T2I models demonstrate the capacity to produce high-quality images, they require appropriate textural descriptions and focus on generating single-character images. Hence, we first introduce the LLM-based character designer, to generate the structured character-oriented prompt book to encode user intentions and descriptions of multiple characters. The details of the user-inputted prompt and the prompt book can be seen in the Appendix.

Formally, the proposed prompt book $P$ can be formulated as, 
\begin{equation}
    P = \{p_{bg},~ p_{neg},~ (ID_i, p_{i}, b_{i})_{i=1}^k\},
\end{equation} 
where $p_{bg}$ is the background prompt, $p_{neg}$ is the negative prompt, and $k$ is the number of characters existing in the current interaction turn. The triple $(ID_i, p_{i}, b_{i})$ represents the unique ID, the prompt, and the layout (bounding box) of the $i$-th character, respectively. With the unique ID assigned by the LLM, we can distinguish and track different characters in multiple turns effectively. This also allows us to edit images of target characters, even if they exhibit significant changes in appearance. Moreover, we can exploit spatial relationship priors from the LLM through the character layout, which helps to arrange multiple characters in the synthesized image reasonably and improve semantic consistency.   


\subsection{Character Image Manager}
\label{sec:Local image generation}

The prompt book serves to reify multi-turn dialogues with users into a more sustainable format. However,  employing the prompt book in T2I models directly still poses challenges in attaining semantic and contextual coherence. Thus, we devised the character image manager, which (i) generates on-stage character images according to the prompt book, and (ii) exploits guidance to ensure semantic and contextual consistency in generated images.

\textbf{On-stage character image generation}. For each character, we consider two types of images, namely, a reference image that is used to maintain contextual consistency throughout the entire interaction process, and an on-stage image that represents the character in the current stage. Specifically, for an arbitrary character with $ID_i$, let $I_{ref}^{(ID_i)}$ and $I_{on}^{(ID_i)}$ denote its reference image and on-stage image, respectively. We generate $I_{on}^{(ID_i)}$ with an off-the-self diffusion-based T2I model $\mathcal{M}$ as follows:

\begin{align}
I_{on}^{(ID_i)} = 
\begin{cases}
    \mathcal{M}(p_i,~ \mathcal{A}(I_{ref}^{(ID_i)})), & \text{if } I_{ref}^{(ID_i)} \in D, \\ \\
    \mathcal{M}(p_i), & \text{if } I_{ref}^{(ID_i)} \notin D,
\end{cases}
\end{align}
where $D$ is a list maintaining the reference images of all characters, i.e.,
\begin{equation}
    D = \{I_{ref}^{(ID_1)}, ..., I_{ref}^{(ID_k)}\}.
\end{equation}
$\mathcal{A}$ is an adapter that injects the reference image into $\mathcal{M}$. With $\mathcal{A}$, we encourage the generated image to exhibit similar characteristics to the reference image and achieve contextual consistency. We adopt a pre-trained IP-Adapter~\cite{ipadapter} to implement $\mathcal{A}$. For a character that appears for the first time, we generate both its reference image and on-stage image via $\mathcal{M}$ directly. 

\begin{figure}[tb]
  \centering
  \includegraphics[width=\textwidth]{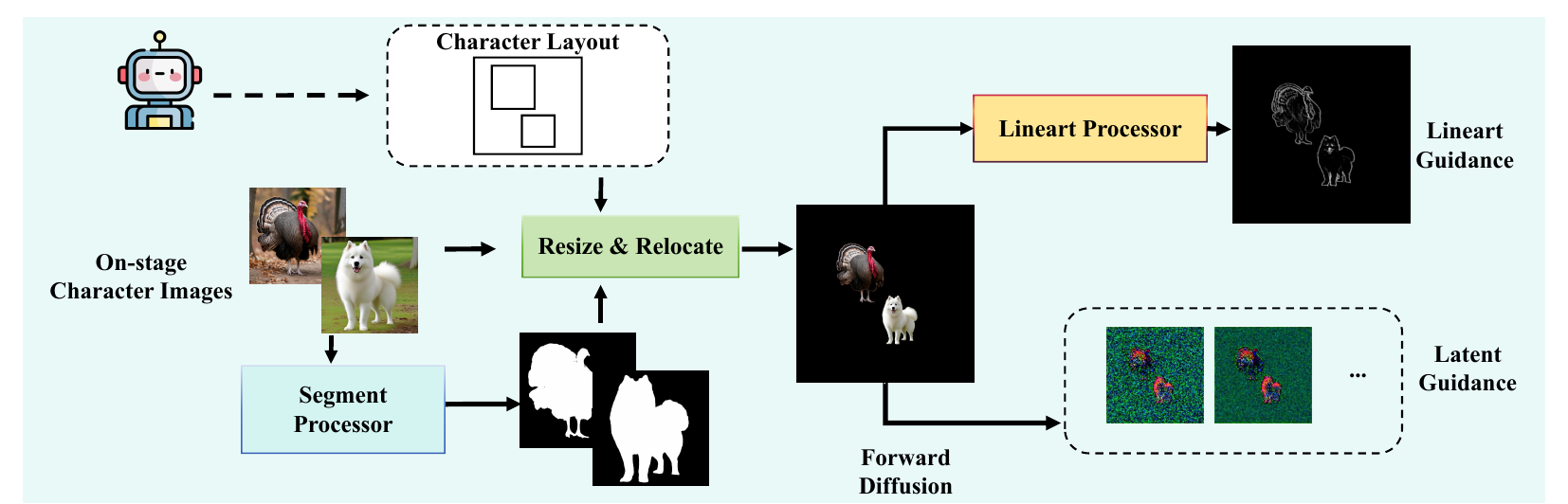}
  \caption{The proposed guidance extractor. It first extracts subjects from character images and rearranges them into the same image according to the layout. Then the lineart guidance and the latent guidance for subsequent image generation are obtained via a lineart processer and the forward diffusion process, respectively.}
  \label{fig:model_detail}
    \vspace{-2em}
\end{figure}

\textbf{Guidance extraction}. To consolidate all on-stage images into one that is coherent with the layouts provided by the prompt book and suitable for the subsequent generation, we propose to extract two types of guidance, as shown in Figure \ref{fig:model_detail}. To begin with, we construct a segment processor comprising an open-vocabulary detection model \cite{GroundingDino} and a segmentation model \cite{SAM} to detect characters and remove backgrounds from the on-stage images. We then resize all on-stage images and merge them into the same one according to the character layouts, yielding a middle-state image $I_{mid}$ with a blank background. 

We consider extracting the latent guidance and the lineart guidance from $I_{mid}$. The former type of guidance is constructed by applying the forward diffusion process of $\mathcal{M}$ on $I_{mid}$, so that we can project $I_{mid}$ into the latent space of $\mathcal{M}$, complete and revise it via the reverse diffusion process. Particularly, the latent guidance $G_{latent}$ is generated as follows,
\begin{equation}
    G_{latent} = \{g_{l, t} | t=0,...,T-1\}, ~ g_{l, t} = \mathcal{FD}(I_{mid}, t),
\end{equation}
where $\mathcal{FD}$ is the forward diffusion process of $T$ steps and $g_{l, t}$ represents the latent guidance at the $t$-th step. In this way, all generated images are from the same latent space. Hence, with $G_{latent}$, we can maintain the characteristics of on-stage images and improve contextual consistency. 

In addition, to attain more precise manipulation of character layout and image details, a lineart processor \cite{controlnet} is employed to extract lineart guidance $G_{lineart}$. This method, as opposed to the direct utilization of bounding boxes like in conventional methods \cite{he2021context,zheng2023layoutdiffusion}, provides stronger constraints on character positions. Moreover, this facilitates the preservation of finer edges and structural details, thereby contributing to the production of high-quality and consistent images.

\subsection{Character-Guided Generator}
\label{sec:Global image generation}

Given the prompt book $P$ and two types of guidance, our devised character-guided generator is responsible for synthesizing the final image for each turn. Specifically, we first concatenate the character prompts and the background prompt in $P$ to obtain a global prompt $p_{global}$.  $p_{global}$ contains the semantic information of the entire image, which we inject into the cross-attention modules of $\mathcal{M}$ along with the guidance information to control image generation. The negative prompt is utilized following the common practice of diffusion models \cite{sd}, and here we focus on exploiting lineart guidance and latent guidance.

\textbf{Lineart guidance}. In each U-Net block of $\mathcal{M}$, we follow \cite{controlnet} to encode the lineart guidance $G_{lineart}$ and inject it into the denoising process. As mentioned above, this strengthens the layout constraint and incorporates more details into synthesized characters. 

\textbf{Latent guidance.} While the global prompt $p_{global}$ and the lineart guidance $G_{lineart}$ are important for establishing semantic consistency, they are insufficient in guaranteeing contextual consistency. 
Therefore, we utilize the latent guidance $G_{latent}$ to enhance the consistency of a character across multiple turns. To this end, let $z_t$ represent the denoised latent code at the $t$-th time step, we incorporate $G_{latent}$ into $z_t$ as follows,
\begin{align}
\hat{z}_t = 
\begin{cases}
    z_t \odot (1-M) + g_{l, t} \odot M, & \text{if } t \geq rT, \\
    z_t,  & \text{if } t < rT, \\
\end{cases}
\end{align}
\begin{equation}
    M = m^{(ID_1)} \oplus m^{(ID_2)}, ..., \oplus m^{(ID_k)},
\end{equation}
where $\odot$ denotes the element-wise product and $\oplus$ denotes the OR operator. $m^{(ID_i)} \in \{0, 1\}^{H \times W}$ denotes the $H \times W$-sized segmentation mask of the $i$-th character obtained by the segment processor. $r$ is the radio to control the number of steps that are required to input latent guidance. We retain a portion of the denoising steps without replacement by $r$, so that we can better balance the integration between the characters and the background in the final image. The augmented $\hat{z}_t$ will be processed as the input of the next denoising step. With $G_{latent}$, reference images and on-stage images can be merged effectively to guide the generation of the final image.

\section{Experiments}

\subsection{Setup}
\subsubsection{CMIGBench}
As shown in Table~\ref{tab:Benchmark comparison}, existing benchmarks have certain limitations for multi-turn T2I generation. Primarily, they fail to comprehensively consider story generation and multi-turn editing, focusing instead on specific types of multi-turn generation. Secondly, some benchmarks disregard contextual consistency, while others fail to involve multi-turn instructions. In addition, most multi-turn benchmarks pre-define the images of characters, requiring models to undergo fine-tuning to generate specific forms of characters to achieve contextual consistency, which is hard to meet the demands of real-world applications.

Therefore, we propose a consistent multi-turn
image generation benchmark (\ourdataset). This benchmark focuses on two common types found in multi-turn image generation: story generation and multi-turn editing, comprising 8000 multi-turn scripted dialogues (4000 for each task) with each consisting of 4 turns of natural language instructions. This allows us to evaluate the semantic consistency and contextual consistency of different models in multi-turn image generation in a zero-shot manner.

CMIGBench is constructed automatically through GPT-4 \cite{gpt4} and then manually curated. To ensure the diversity of dialogues, we first randomly select the characters and a background scene for each dialogue from a pre-defined pool. Then, GPT-4 is prompted to generate image-generation instructions for 4 turns and a ground-truth prompt book. We encourage GPT-4 to use unambiguous references to make the instructions more human-like. In the multi-turn editing task, to avoid ambiguity, we have specifically fixed the editing type for each turn, namely, editing spatial locations, attributes, negative prompts, and numbers of characters. This facilitates the generation of high-quality dialogues and enables the evaluation of alignment for each specific edit instruction.

\begin{table}[t!]
    \caption{Comparative analysis of benchmarks on diversity and functionality metrics. "Char. Predefined" means characters are pre-defined.}
    \label{tab:Benchmark comparison}
    \centering
\resizebox{1\textwidth}{!}{
    \begin{tabular}{lcc|c|c|c|c}
        \toprule
        Benchmarks & Vocabulary Diversity & Test Prompts & Char. Pre-defined & Multi-turn & Story & Edit \\
        \midrule
        LMD\cite{lian2023llm}  & 10 objects, 11 colors, 4 spatial & 400 & \ding{55} & \ding{55} & \ding{55} & \checkmark\\
         ConsiStory\cite{tewel2024training} & 31 nouns & 500 & \ding{55} & \ding{55} &\checkmark & \ding{55} \\
        CLEVR-SV\cite{li2019storygan} & 8 colors, 3 shapes, 2 sizes & 12000 & \checkmark & \checkmark & \ding{55} & \checkmark \\
        CoDraw\cite{Wang_2024_WACV} & 58 nouns & 4250 & \checkmark & \checkmark & \ding{55} & \checkmark \\
        PororoSV\cite{li2019storygan} & 13 characters &  11680 & \checkmark & \checkmark &\checkmark & \ding{55} \\
        FlintstonesSV\cite{gupta2018imagine} & 3897 characters  & 2518 & \checkmark & \checkmark &\checkmark & \ding{55}\\
        \textbf{\ourdataset} & 157 nouns, 16 colors, 6 spatial & 32000 & \ding{55} & \checkmark & \checkmark & \checkmark \\
        \bottomrule
    \end{tabular}}
\end{table}

\subsubsection{Evaluation metrics}
To evaluate the contextual consistency, in the absence of pre-defined character images, we utilize character reference images for evaluation. For every dialogue, we use Grounding DINO to identify character regions in the final image of each turn. Each character's initial appearance is set as its reference. For detected character regions and their corresponding reference image, the assessment is conducted by calculating: (1) the average character-to-character similarity (aCCS) through the CLIP image-image similarity method, (2) the average Frechet Inception Distance (aFID). 

Moreover, for evaluating semantic consistency, we leverage CLIP to determine the average text-image similarity (aTIS) for the final image of each turn and the associated global prompt. Additionally, to measure editing alignment accuracy in the context of editing tasks, we follow \cite{LMD} to compute metrics w.r.t. our defined 4 editing types ("spatial", "attribute", "negative", and "numeracy").

\subsubsection{Inference Details}
For LLM Character Designer, we use GPT-4 for prompt book generation while also evaluating the ability of other LLMs (\emph{i.e.} GPT-3.5 and Gemini Pro~\cite{team2023gemini}) to generate prompt books, which determines the upper limit of our model's generation quality. For image generation, Stable Diffusion version 1.5 and version XL are selected as the basic generation models for TheaterGen. 
For the Segment Processor, we utilize the GroundingDINO-T with a box-threshold setting of 0.5 and a text-threshold setting of 0.25 and SAM \cite{c} to predict mask. The $r$ to control the injection of latent guidance is set to 0.1.

\begin{table}[!t]
  \caption{Model performances on contextual and semantic consistency metrics.}
  \label{tab:Story generation}
  \centering
\resizebox{0.9\textwidth}{!}{
\begin{tabular}{cc|cc|cc|cc|cc|cc}
\toprule
\multirow{4}{*}{Diffsion version} & \multicolumn{1}{c}{\multirow{4}{*}{Model}} & \multicolumn{10}{c}{Metrics} \\
 & \multicolumn{1}{c}{} & \multicolumn{6}{c}{Contexual consistency} & \multicolumn{4}{c}{Sementic consistency}  \\
 \cmidrule(lr){3-8} \cmidrule(lr){9-12}
 & \multicolumn{1}{c}{} & \multicolumn{2}{c}{aFID$\downarrow$} & \multicolumn{2}{c}{aCCS(\%)$\uparrow$} & \multicolumn{2}{c}{Human} & \multicolumn{2}{c}{aTIS(\%)$\uparrow$} & \multicolumn{2}{c}{Human}  \\
 && Story & Editing & Story & Editing & Story & Editing & Story & Editing & Story & Editing \\
\midrule
\multirow{4}{*}{SD1.5} & Mini DALLE·3 & 451.59&443.71 & 54.11&52.51 & 1.72&1.86 & 29.81&28.13 & 2.48&2.12  \\
 & MiniGPT-5 & 528.3&480.7 & 43.09&44.08 & 1.24&1.08 & 24.93&22.95 & 1.94&1.41  \\
 & SEED-LLaMA & 316&357.55 & 64.78&59.83 & 1.52&1.42 & 26.41&25.18 & 2.11&2.26  \\
 & \textbf{Ours} &  \textbf{252.31}&\textbf{240.32} & \textbf{78}&\textbf{84.31} & \textbf{2.64}&\textbf{3.63} & \textbf{31.52}&\textbf{29.67} & \textbf{3.06}&\textbf{3.55}  \\
 \midrule
\multirow{2}{*}{SDXL} & Mini DALLE·3 & 286.21 &402.21 & 67.59& 54.4 & 2.72&1.69  & 32.77& 29.86 & 3.22&2.47  \\
 & \textbf{Ours} & \textbf{209.45}&\textbf{222.56}  & \textbf{81.05}& \textbf{93.52} & \textbf{3.18}& \textbf{4.18} & \textbf{38.91}&\textbf{37.72}  & \textbf{3.24}& \textbf{4.11}  \\
 \bottomrule
\end{tabular}}
\end{table}

\begin{table}[!t]
  \caption{Model performances on alignment accuracy metrics.}
  \label{tab:Editing}
  \centering
\resizebox{0.9\textwidth}{!}{
\begin{tabular}{cc|c|c|c|c|c}
\toprule
\multirow{2}{*}{Diffsion version} & \multicolumn{1}{c}{\multirow{2}{*}{Model}} & \multicolumn{5}{c}{Alignment Accuracy $\uparrow$} \\
  \cmidrule(lr){3-7} 
 &  & Spatial(\%) & Attribute(\%) & Negative(\%) & Numeracy(\%) & Human \\
 \midrule
\multirow{4}{*}{SD1.5} & Mini DALLE·3 & 38.32 & 64.72 & 85.04 & 7.83 & 1.87  \\
 & MiniGPT-5 &  34.17 & 49.6 & 87.68 & 3.65 & 1.12  \\
 & SEED-LLaMA &  44.99 & 71.92 & 72.05 & 9.51 & 1.35  \\
 & \textbf{Ours} & \textbf{87.27} & \textbf{88.21} & \textbf{89.31} & \textbf{47.74} & \textbf{3.24}  \\\hline
\multirow{2}{*}{SDXL} & Mini DALLE·3 & 40.8 & 67.6 & 87.33 & 5.83 & 1.71 \\
 & \textbf{Ours} & \textbf{87.33} & \textbf{90.66} & \textbf{94.92} & \textbf{70.67} & \textbf{3.41}  \\
 
\bottomrule
\end{tabular}}
  \vspace{-1em}
\end{table}

\subsection{Comparison with SOTA methods}
\label{sec: Qualitative analysis}

We do a comprehensive evaluation of TheaterGen with the chosen baseline models on CMIGBench.

As shown in Table~\ref{tab:Story generation} and Table~\ref{tab:Editing}, our approach surpasses other comparative models across all metrics in both the story generation task and the multi-turn image editing task. Specifically, our model showcases a substantial enhancement of 27\% and 21\% respectively in the aFID and aCCS metrics about semantic consistency, when contrasted with the previous SOTA model Mini DALLE·3 with diffusion version XL. Additionally, in terms of contextual consistency, our model demonstrates an improvement of 19\% in the aTIS metrics compared to the previous SOTA model. These results substantiate the notion that our model exhibits a heightened capacity to generate images on multi-turn dialog that upholds both semantic and contextual consistency.

For the alignment accuracy, Table~\ref{tab:Editing} demonstrates that our model not only performs well in maintaining dialogue consistency but also outperforms other baselines in terms of completion for each editing type. A significant improvement in our model's performance is observed, particularly in cases involving spatial relationships and quantities. This is mainly attributed to the layout capability of the LLM.

\vspace{-1em}
\subsection{Human evaluation}
\begin{figure*}[!t]
  \centering
  \includegraphics[width=\textwidth]{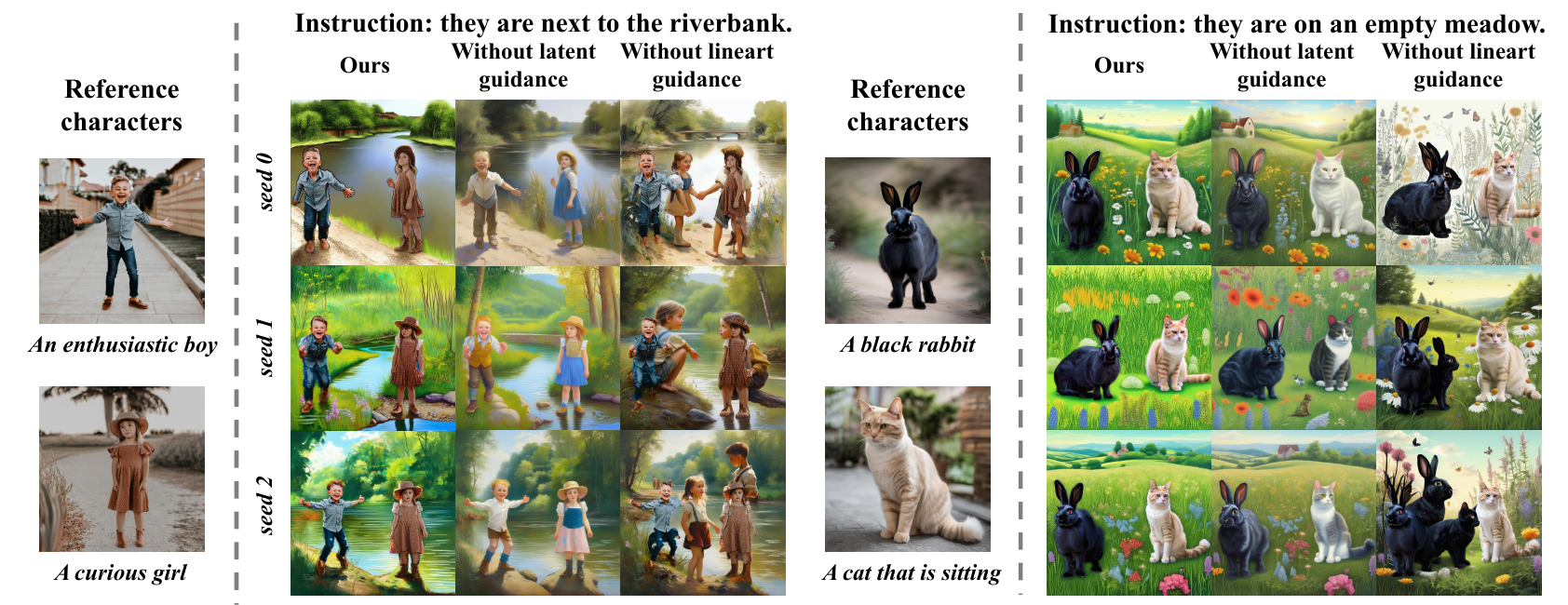}
  \caption{Ablation study on the effects of lineart and latent guidance.}
  \label{fig:ablation}
\end{figure*}

To demonstrate the effectiveness of our evaluation process and metrics, as well as to provide an intuitive comparison of model performance, we conduct a user study. We randomly select 50 dialogues (25 for each task) from \ourdataset, resulting in 200 images generated per model. 10 volunteers are invited to rate the semantic consistency and contextual consistency of these images. 

From the human evaluation results in Table~\ref{tab:Story generation} and Table~\ref{tab:Editing}, we can observe that human ratings of our method are higher than those of other methods. This further validates the effectiveness of our method.

Besides, we compute Kendall's correlation coefficient $\tau$, Spearman's correlation coefficient $\rho$, and Pearson correlation coefficient $\sigma$ between our evaluation metrics and the results from human evaluation. From Table~\ref{tab:Correlation}, it can be observed that all the metrics demonstrate strong correlations, indicating that the proposed metrics are minimally affected by the errors introduced by the detection model and can effectively reflect the quality of multi-turn image generation. \\

\begin{table}[!tb]
  \caption{Correlation between machine metrics and human evaluations. The correlation of aFID is expected to be close to -1, while the correlation of the other metrics is expected to be close to 1.}
  \label{tab:Correlation}
  \centering
\resizebox{0.7\textwidth}{!}{
    \begin{tabular}{cccccccc}
    \toprule
    Metrics & aFID & aCCS & aTIS & Spatial & Attribute & Negative & Numeracy \\
    \midrule
    $\tau$ & -0.349 & 0.287 & 0.439 & 0.633 & 0.604 & 0.600 & 0.687  \\
    $\rho$ & -0.528 & 0.381 & 0.611 & 0.678 & 0.629 & 0.610 & 0.733  \\
    $\sigma$    & -0.401 & 0.426 & 0.641 & 0.699 & 0.662 & 0.904 & 0.731 \\
    \bottomrule
\end{tabular}}
\end{table}

\makeatletter\def\@captype{table}\makeatother 
\begin{minipage}{.5\linewidth}
    \caption{Ablation results of two guidance approaches with consistency metrics.}
    \centering
    \resizebox{1\textwidth}{14mm}{
    \renewcommand{\arraystretch}{1.25}
    \begin{tabular}{c|cc|ccc}
         Task & Lineart& Latent& {aFID$\downarrow$}& {aCCS(\%)$\uparrow$}& {aTIS(\%)$\uparrow$} \\\hline
         \multirow{2}{*}{Story} & \checkmark & \checkmark &\textbf{252.31} &\textbf{78} &\textbf{31.52} \\
         & \checkmark &  &281.17 & 76.96&31.33 \\
         &  & \checkmark &306.24 &68.45 &31.14 \\\hline
         \multirow{2}{*}{Editing} & \checkmark & \checkmark &\textbf{240.32} &\textbf{84.31} &\textbf{29.67} \\
         & \checkmark &  &288.25 &77.66 &28.46 \\
         &  & \checkmark &300.54 &78.07 &28.85 \\\hline
    \end{tabular}
    }
    \label{tab:ablation guidance story}
\end{minipage}
\makeatletter\def\@captype{table}\makeatother 
\begin{minipage}{.5\linewidth}
    
    \caption{Ablation results of two guidance approaches with alignment accuracy metrics.}
    \centering
    
    \resizebox{1\textwidth}{12mm}{
    \renewcommand{\arraystretch}{2.0}
    
    \begin{tabular}{cc|cccc}
        Lineart& Latent&{Spatial(\%)}&{Attribute(\%)}&{Negative(\%)}&{Numeracy(\%)}\\\hline
         \checkmark & \checkmark & \large \textbf{87.27} & \large \textbf{88.21} & \large \textbf{89.31} & \large \textbf{47.74} \\
        \checkmark &  & \large 81.48 & \large 88.76 & \large 79.78 & \large 42.32\\
        &\checkmark  &\large 58.36 & \large 67.91 & \large 75.52 & \large 14.93\\\hline
    \end{tabular}
    }
    \label{tab:ablation guidance editing}
\end{minipage}

\subsection{Ablation experiment}

\subsubsection{Latent guidance}
As shown in Table~\ref{tab:ablation guidance story} and Table~\ref{tab:ablation guidance editing}, we conducted ablation experiments on latent guidance, revealing its importance for multi-turn image generation. In the comparison between the second and first columns of Figure~\ref{fig:ablation}, we can more intuitively see the role of latent guidance. It can be seen that without latent guidance, which means that we do not introduce the image features of the reference images into the diffusion model,   the character in the generated images will be inconsistent with those in the reference images, causing the model to fail to maintain contextual consistency across multi-turn generations.

\subsubsection{Lineart guidance}

As shown in Table~\ref{tab:ablation guidance story} and Table~\ref{tab:ablation guidance editing}, we also conducted ablation study on the role of lineart guidance. It can find that  after removing $G_{lineart}$, all metrics of the model will experience a significant decrease, verifying the effectiveness of $G_{lineart}$.  In Figure~\ref{fig:ablation}, we can observe that in the absence of lineart guidance, the model struggles to handle details such as the positions and quantities of characters. Additionally, the model fails to effectively process latent guidance, leading to conflicts between characters and background. This indicates that our two types of guidance complement each other.



\captionsetup[figure]{skip=0pt}
\begin{figure*}[!t]
  \centering
  \includegraphics[width=0.95\textwidth]{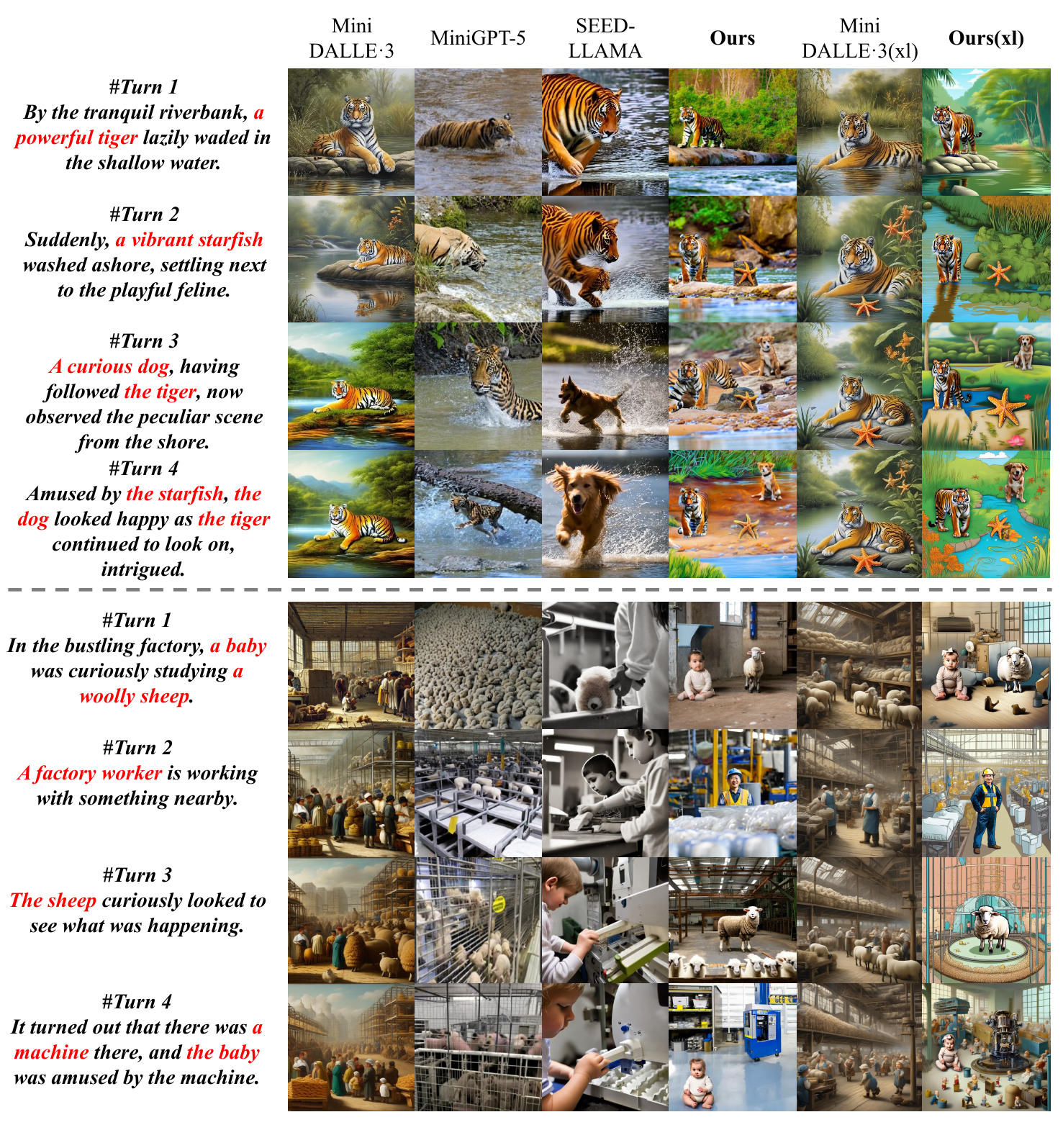}
  \caption{Visualization results of story generation.}
  \label{fig:main result 1}
\end{figure*}

\vspace{\baselineskip}
\makeatletter\def\@captype{table}\makeatother 
\begin{minipage}{.45\linewidth}
\caption{Ablation results of different LLMs with consistency metrics.}
    \centering
    \resizebox{1\textwidth}{12mm}{
    \renewcommand{\arraystretch}{1.0}
    \begin{tabular}{c|c|cc}
         LLM & Task& $ACC_{id}(\%)$ & $ACC_{box}(\%)$ \\\hline
         GPT-3.5&\multirow{3}{*}{Story} &96.21 &82.68 \\
         Gemini-pro& & 95.33 &95.10 \\
         GPT-4&  & 99.73&92.48 \\\hline
         GPT-3.5&\multirow{3}{*}{Editing} &100 & 96.97\\
         Gemini-pro& & 99.5&96.97 \\
         GPT-4&  &100 &99.63 \\\hline
    \end{tabular}
    }
    \label{tab:ablation llm story}
\end{minipage}
\makeatletter\def\@captype{table}\makeatother 
\begin{minipage}{.5\linewidth}
    \caption{Ablation results of different LLMs with alignment accuracy metrics.} 
    \centering
    \resizebox{1\textwidth}{12mm}{
    \renewcommand{\arraystretch}{2.4}
        \begin{tabular}{c|cccc}
        \Large  LLM 
        &  {Spatial(\%)}
        &  {Attribute(\%)}
        &  {Negative(\%)}
        &  {Numeracy(\%)} \\
        \hline
         \Large GPT-3.5 &  \Large 90 
        &  \Large 84.85 & \Large 95.96 & \Large 96.97 \\
         \Large Gemini-pro & \Large 91.92 & \Large 96.97 & \Large 96.97 & \Large 100 \\
         \Large GPT-4 
        & \Large 100 
        & \Large 100 
        & \Large 100 
        & \Large 100 \\
        \bottomrule
        \end{tabular}
    }
    \label{tab:ablation llm editing}
\end{minipage}

\subsubsection{LLM Character Designer}

We also evaluate the recent LLM models' ability to generate proper prompt books. As shown in Table~\ref{tab:ablation llm story} and Table~\ref{tab:ablation llm editing}, their generation capabilities for editing instructions are slightly different, yet the overall quality is satisfying. Among them, GPT-4 exhibits the best average performance. We evaluated the accuracy of LLM in maintaining the corresponding character ID as $ACC_{id}$ and the accuracy of generating bounding boxes that contain the required main character as $ACC_{box}$. The results in Table~\ref{tab:Editing} of the prompt book show that our selected LLMs assign proper character IDs and generate layouts that are consistent with the captions in most cases.

\vspace{-1em}
\subsection{Visualization}

\begin{figure*}[!t]
  \centering
  \includegraphics[width=0.95\textwidth]{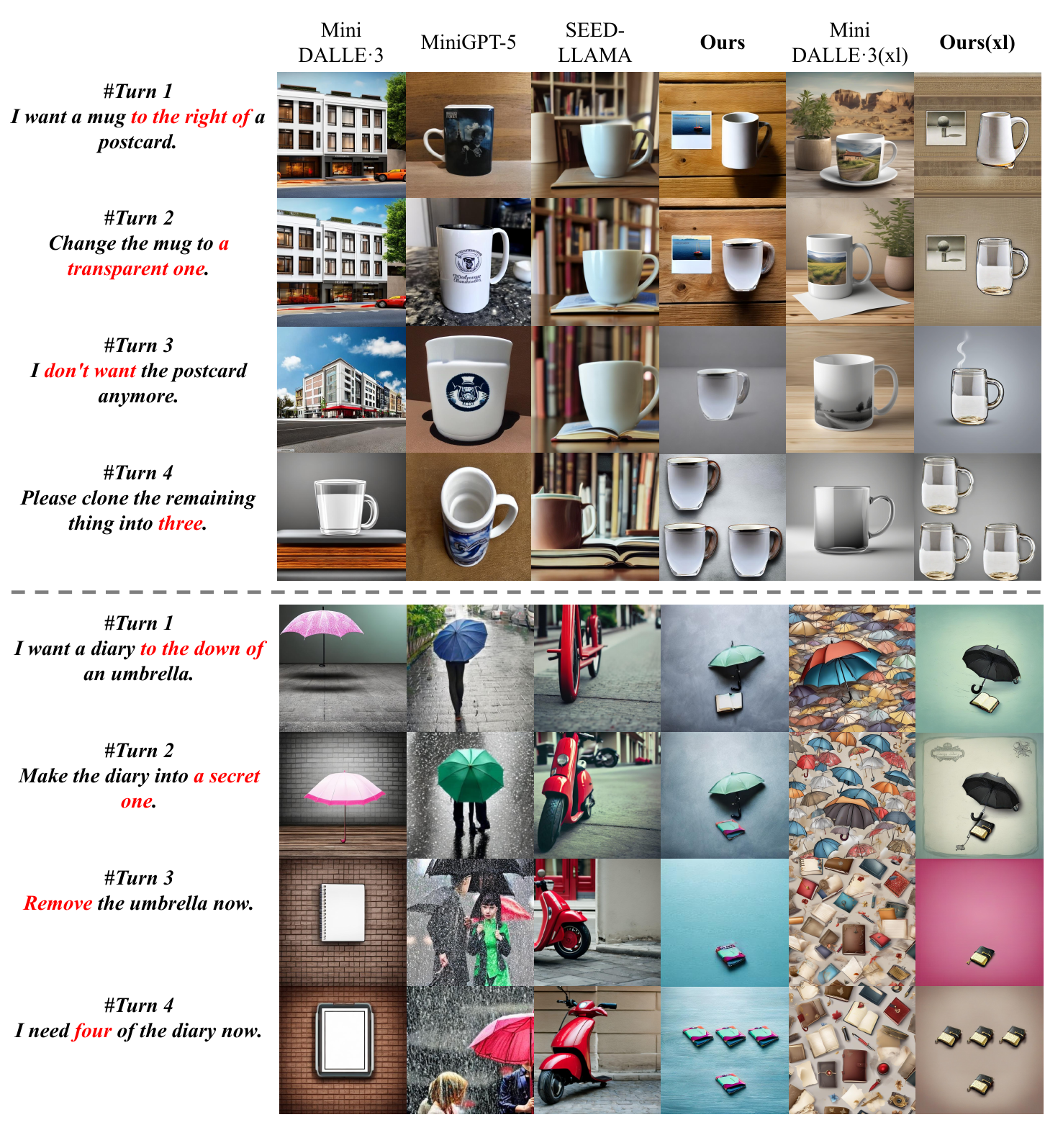}
  \caption{Visualization results of multi-turn editing.}
  \label{fig:main result 2}
\end{figure*}

We visualize the generation results of the chosen models. In Figure \ref{fig:main result 1} and Figure \ref{fig:main result 2}, we observe that the images generated by TheaterGen perform well in both semantic and contextual consistency. This suggest \ourmodel~can handle complex referential issues and maintain discernible features for the main characters. More results can be seen in Appendix. 

\section{Discussion and Conclusion}
\textbf{Conclusion}. In this paper, we propose \ourmodel, a training-free framework integrating LLMs with T2I models for multi-turn image generation. By leveraging a large language model as a "Screenwriter", \ourmodel manages a prompt book with character prompts and layouts. It then introduces a character image manager, which customizes the image generation for each character and arranges the "Rehearsal" for the entire stage scene, generating latent guidance and lineart guidance. To generate the ``Final Performance" image, \ourmodel utilizes the prompt book and guidance information to incorporate customized content effectively, and eventually addresses the challenges of maintaining semantic and contextual consistency in multi-turn image generation. \\

\textbf{Limitations}. Despite customizing the generation for each character, the randomness in existing pre-trained image adapters might result in contextual inconsistency. Additionally, limited by the generation capability, T2I models might cause semantic inconsistency. In the future, more power T2I models will be incorporated into TheaterGen to address the above issues.

\newpage
\appendix

\title{Appendix Materials of TheaterGen} 

\section{Prompt book Generation Details}

We utilize an LLM to translate human instructions into a prompt book. Our approach incorporates the user instruction into a template and requests the LLM to generate the rest of the content, yielding the following prompt:

\begin{tcolorbox}
\textbf{<Task Description>}:
Your task is to generate the bounding boxes for 
the objects mentioned in the caption and number them, 
along with a background prompt describing the scene. 
The number indicates the number of mentioned objects.

\textbf{<Supporting details>}:
The images are of size 512 x 512. The top-left corner has coordinates [0, 0]. The bottom-right corner has coordinates [512, 512]. Pay attention to ensure the generated layout size is appropriate and not too small. The bounding boxes should not overlap or go beyond the image boundaries. Each bounding box should be in the format of ( an object with an article a or an and a modifier, [ top - left x coordinate, top - left y coordinate, box width, box height ], object number) and the content in '' must be the singular form of a or an plus an adjective plus a countable noun and should not include more than one object. You should completely avoid situations where the bounding boxes of objects overlap, so you can make the bounding boxes between different objects have an appropriate distance and need design reasonable x, y, box width, and box height. Do not put objects that are already provided in the bounding boxes into the background prompt. When describing the same picture, different objects should not have the same numbers. If you think the description refers to the same objects in the previous conservation, its number should be consistent with the previous conservation, such as a green apple and a red apple. Do not include non-existing or excluded objects in the background prompt. Use "a realistic scene" as the background prompt if no background is given in the prompt. If needed, you can make reasonable guesses.

\textbf{<Examples>}: several examples of multi-turn layout generation.

\textbf{<History dialogue>}: The history dialogue

\textbf{<Current instruction>}: Current prompt by human.
\end{tcolorbox}

Once the above prompt is obtained, the LLM is expected to generate the prompt book in the following format:

\begin{tcolorbox}
\textbf{<Characters>}: [(an object with an article a or an and a modifier, [ top-left x coordinate , top-left y coordinate , box width, box height ], object number)] \\
\textbf{<Background prompt>}: The instruction of background.\\
\textbf{<Negative prompt>}: The instruction of which object is not supposed to appear. \\
\end{tcolorbox}

\section{Model Deployment Details}
\subsection{Models Input}
As our selected baselines have different designs, their inputs vary slightly. In our model, we utilize the prompt book as input directly. For MiniGPT-5 and SEED-LLaMA, we construct a combined input that consists of the dialogue history, the current instruction, and their output w.r.t the current instruction. As for Mini DALLE·3, it employs an LLM to transform the instruction into commands for image generation. In this case, we uniformly use the global prompt $p_{global}$ in our prompt book and incorporate the "</edit>" tag provided by Mini DALLE·3 as its input. This enables Mini DALLE·3 to consider the historical results during the current generation turn.

\subsection{Inference Time}
In our experiments, the diffusion time steps for SD1.5 and SDXL are set to 50 and 30. When applying SD1.5, the baselines take about 10 to 15 seconds to complete a dialogue (4 turns), while our TheaterGen requires about 35 to 40 seconds. This is reasonable, as we generate on-stage character images in each turn. And such a strategy is worthwhile, as our experimental results validate that it improves image quality and consistency significantly. Moreover, with more powerful diffusion models, the gap in computational costs tends to be marginal. For example, with SDXL, Mini DALLE·3 takes approximately 36 seconds for a dialogue, while while TheaterGen requires around 86 seconds.


\section{Dataset Construction Details}

\subsection{Character Pools}

We create separate pools from 50 common fruits, 100 objects, 7 humans, and 50 scenes, respectively, to ensure the diversity of characters. Our selected characters are listed as follows:

\begin{tcolorbox}
\textbf{fruit pool} = [`apple', `banana', `orange', `pear', `grape', `strawberry', `watermelon', `cantaloupe', `cherry', `peach', `plum', `kiwi', `pineapple', `mango', `lemon', `pomegranate', `lychee', `papaya', `blueberry', `raspberry', `blackberry', `apricot', `nectarine', `grapefruit', `lime', `tangerine', `fig', `guava', `persimmon', `dragon fruit', `passion fruit', `star fruit', `jackfruit', `durian', `avocado', `cranberry', `mulberry', `quince', `kumquat', `date', `coconut', `olive', `honeydew', `currant', `gooseberry', `boysenberry', `elderberry', `pummelo', `cloudberry', `jujube']

\textbf{object pool} = [`cup', `book', `pen', `car', `glasses', `chair', `phone', `laptop', `key', `wallet', `watch', `shoe', `sock', `hat', `shirt', `pants', `dress', `coat', `pillow', `towel', `soap', `toothbrush', `broom', `camera', `clock', `lamp', `scissors', `notebook', `backpack', `umbrella', `bowl', `plate', `fork', `knife', `spoon', `toothpaste', `brush', `balloon', `battery', `blanket', `headphones', `paper', `envelope', `stamp', `card', `toy', `plant', `vase', `necklace', `belt', `bicycle', `mirror', `comb', `bed', `cushion', `diary', `teapot', `kettle', `mug', `earphones', `charger', `remote', `skateboard', `surfboard', `ball', `glove', `sunglasses', `tie', `scarf', `calendar', `flashlight', `tent', `grill', `hammock', `bucket', `teacup', `saucer', `colander', `pitcher', `whisk', `spatula', `ladle', `cutting board', `measuring cup', `cookie sheet', `blender', `mixer', `oven mitt', `apron', `trash can', `recycling bin', `suitcase', `map',`postcard', `album', `puzzle',   mosquito,  tissue,  coin,  eraser]

\textbf{animal pool} = [`dog', `cat', `rabbit', `horse', `cow', `sheep', `pig', `goat', `chicken', `duck', `turkey', `deer', `elephant', `tiger', `lion', `bear', `monkey', `kangaroo', `panda', `zebra', `penguin', `frog', `toad', `lizard', `snake', `turtle', `crocodile', `parrot', `sparrow', `eagle', `bat', `snail', `crab', `lobster', `starfish']

\textbf{human pool} = [`baby', `boy', `girl', `man', `woman', `old man', `old woman']

\textbf{background pool} = [`park', `beach', `zoo', `forest', `school', `playground', `city street', `suburban neighborhood', `mountain range', `countryside', `farm', `garden', `riverbank', `desert', `ocean', `island', `downtown skyline', `harbor', `lake', `meadow', `campsite', `hiking trail', `waterfall', `bridge', `marketplace', `amusement park', `aquarium', `library', `museum', `art gallery', `cafe', `restaurant', `bakery', `bookstore', `gym', `stadium', `arena', `theater', `cinema', `concert hall', `classroom', `office', `warehouse', `factory', `hospital', `clinic', `police station', `fire station', `train station', `airport']
\end{tcolorbox}

\subsection{Prompt for Story Generation}

After randomly selecting characters from the pools, we use the following format of prompts as input to GPT-4 for generating story generation dialogues simulating human conversations, along with the ground truth prompt book.

\begin{tcolorbox}
\textbf{<characters string>}: Randomly selected characters from human and animal pool

\textbf{<scene string>}: Randomly selected background from background pool

\textbf{<prompt>}: "Please give me a story with " + characters string + " as protagonists. The setting background is " + scene string + ". Each story contains four simple and human natural language sentences. Each sentence contains less than 4 objects. All objects are countable nouns and are singular, with no postattributive. If the protagonist has been mentioned in the previous text, subsequent sentences should use pronouns to refer to the protagonist. The protagonist must be something with a precise number of nouns. You should generate the bounding boxes for the objects mentioned in the caption and number them, along with a background prompt describing the scene. The number indicates the number of mentioned objects. The images are of size 512 x 512. The top-left corner has coordinates [0, 0]. The bottom-right corner has coordinates [512, 512]. Pay attention to ensure the generated layout size is appropriate and not too small. The bounding boxes should not overlap or go beyond the image boundaries. Each bounding box should be in the format of ( an object with an article a or an and a modifier, [ top - left x coordinate, top - left y coordinate, box width, box height ], object number) and the content in '' must be the singular form of a or an plus an adjective plus a countable noun and should not include more than one object. You should completely avoid situations where the bounding boxes of objects overlap, so you can make the bounding boxes between different objects have an appropriate distance and need design reasonable x, y, box width, and box height. Do not put objects that are already provided in the bounding boxes into the background prompt. When describing the same picture, different objects should not have the same numbers. If you think the description refers to the same objects in the previous conservation, its number should be consistent with the previous conservation. Do not include non-existing or excluded objects in the background prompt. Use 'a realistic scene' as the background prompt if no background is given in the prompt. If needed, you can make reasonable guesses. Please refer to the example below for the desired format. The following is an example of a story and its layout, and you needn't mimic its sentence structure, you should focus on the format of the following example:...... Now generate the story, the box for each character should be as large as possible."
\end{tcolorbox}

\subsection{Prompt for Multi-turn Editing}

We use the following input to obtain the dataset for multi-turn editing similar to story generation.

\begin{tcolorbox}
\textbf{<characters string>}: Randomly selected characters from object and fruit pools

\textbf{<number pick>}: Randomly selected from ([`two', `three', `four', `five'])

\textbf{<relation pick>}: Randomly selected from ([`to the left of', `to the right of', `to the top of', `to the down of'])

\textbf{<prompt>}: "You need to simulate a series of multi-round editing instructions. You will receive an object list:[]. The [Spacial Round] should focus on spatial relations. The [Attribute round] should focus on attributes. The [Negative Round] should focus on negatives. The [Numeracy Round] should focus on numeracy. If the object has been mentioned in the last text, subsequent sentences should use pronouns to refer to the protagonist. You should generate the bounding boxes for the objects mentioned in the caption and give them an id, along with a background prompt (empty). The images are of size 512 x 512. The top-left corner has coordinates [0, 0]. The bottom-right corner has coordinates [512, 512]. The bounding boxes should not overlap or go beyond the image boundaries. Each bounding box should be in the format of ( an object with an article and a modifier, [ top - left x coordinate, top - left y coordinate, box width, box height ], object number) and the content in '' must be the singular form of a or an plus an adjective plus a countable noun and should not include more than one object. Do not put objects that are already provided in the bounding boxes ito the background prompt. If you think the description refers to the same objects in the previous conservation, its id should be consistent with the previous conservation(even changed attribute or numeracy). Do not include non-existing or excluded objects in the background prompt. The dialogue must include four types of edit rounds, and [indicate the type] should be used at the beginning. Here is an example output...... You must use quantities at [Numeracy Round] as " + number pick + " and positional relationships in [Spacial Round] as " + relation pick + ". Now, generate the following object list:" + characters string.
\end{tcolorbox}

\subsection{Automated \& Human Corrections}
We further validate and rectify the dialogues generated by GPT-4 to yield the final CMIGBench.


First, we streamline the output format by utilizing regular expressions to identify and rectify format discrepancies across various scenarios, enabling us to reconstruct dialogues that are imperfect in form (but are correct in content). This includes differentiating between text and lists, addressing punctuation inconsistencies across languages, and resolving unenclosed brackets.

Secondly, we detect characters that are mistakenly placed in the background and have GPT-4 regenerate them.

Next, we devise a dispersion strategy to refine the overlap of character layouts within individual turns. For layouts where overlap exceeds a predefined threshold $T_o$, we adjust their positions as follows: We first calculate the collective bounding rectangle of the overlapping layout elements. We then disperse the elements from the center of the rectangle with a random distance. This dispersion step is iterated multiple times and if the overlap is still larger than $T_o$ after that, we utilize GPT-4 to regenerate the layout. We set $T_o$ as $25\%$ in this paper. 

Last but not least, we implement a toolkit for manual visualization alongside drag-and-drop functionalities for layout adjustments. This allows for the hands-on verification and refinement of dialogues and eventually ensures a higher level of accuracy and presentation quality of our dataset.

\subsection{Benchmark Examples}

Here we provide the first dialogues of our story generation and multi-turn editing task to show the description format of our benchmark:

\begin{tcolorbox}[title=Story Generation, fonttitle=\bfseries, colback=white]
\texttt{\textbf{"dialogue 1"}: \{ \\
\ \ \textbf{"characters"}: ["sparrow", "lion", "eagle"], \\
\ \ \textbf{"scene"}: ["library"], \\
\ \ \textbf{"turn 1"}: \{ \\
\ \ \ \ \textbf{"caption"}: "In the silent library, a tiny sparrow was fluttering near a shelf.", \\
\ \ \ \ \textbf{"objects"}: [ \\
\ \ \ \ \ \ \ ["a tiny sparrow", [115, 170, 89, 59], 1], \\
\ \ \ \ \ \ \ ["a library shelf", [215, 165, 171, 171], 2]] \\
\ \ \ \ \textbf{"background"}: "A silent library", \\
\ \ \ \ \textbf{"negative"}: "None" \\
\ \ \}, \\
\ \ \textbf{"turn 2"}: \{ \\
\ \ \ \ \textbf{"caption"}: "An attentive lion in one corner was carefully observing the bird and holding its breath.", \\
\ \ \ \ \textbf{"objects"}: [ \\
\ \ \ \ \ \ \ ["an attentive lion", [300, 221, 162, 180], 3], \\
\ \ \ \ \ \ \ ["a tiny sparrow", [40, 101, 89, 59], 1]] \\
\ \ \ \ \textbf{"background"}: "A silent library", \\
\ \ \ \ \textbf{"negative"}: "None" \\
\ \ \}, \\
\ }
\end{tcolorbox}
\begin{tcolorbox}[title=Story Generation (Continued), fonttitle=\bfseries, colback=white]
\texttt{\textbf{"turn 3"}: \{ \\
\ \ \ \ \textbf{"caption"}: "Above them, a vigilant eagle watched the suspenseful scene unfold from the library ceiling.", \\
\ \ \ \ \textbf{"objects"}: [ \\
\ \ \ \ \ \ \ ["a vigilant eagle", [345, 41, 119, 72], 4], \\
\ \ \ \ \ \ \ ["an observing lion", [295, 281, 162, 180], 3], \\
\ \ \ \ \ \ \ ["a sparrow", [45, 171, 89, 59], 1]] \\
\ \ \ \ \textbf{"background"}: "A silent library", \\
\ \ \ \ \textbf{"negative"}: "None" \\
\ \ \}, \\
\ \ \textbf{"turn 4"}: \{ \\
\ \ \ \ \textbf{"caption"}: "The scenario ended peacefully as the eagle, the lion, and the sparrow all resumed their own activities in the vast library.", \\
\ \ \ \ \textbf{"objects"}: [ \\
\ \ \ \ \ \ \ ["an occupying eagle", [335, 41, 119, 72], 4], \\
\ \ \ \ \ \ \ ["a peaceful lion", [285, 281, 162, 180], 3], \\
\ \ \ \ \ \ \ ["a sparrow", [55, 181, 89, 59], 1]] \\
\ \ \ \ \textbf{"background"}: "A vast library", \\
\ \ \ \ \textbf{"negative"}: "None" \\
\ \ \} \\
\}}
\end{tcolorbox}

\begin{tcolorbox}[title=Multi-Turn Editing, fonttitle=\bfseries, colback=white]
\texttt{\textbf{"dialogue 1"}: \{ \\
\ \ \textbf{"characters"}: ["spatula", "pen"], \\
\ \ \textbf{"scene"}: "empty background", \\
\ \ \textbf{"turn 1"}: \{ \\
\ \ \ \ \textbf{"caption"}: "I want a pen down of a spatula", \\
\ \ \ \ \textbf{"objects"}: [ \\
\ \ \ \ \ \ \ ["a pen", [97, 235, 162, 222], 1], \\
\ \ \ \ \ \ \ ["a spatula", [217, 55, 198, 232], 2]] \\
\ \ \ \ \textbf{"background"}: "empty background", \\
\ \ \ \ \textbf{"negative"}: "None" \\
\ \ \}, \\
\ }
\end{tcolorbox}
\begin{tcolorbox}[title=Multi-Turn Editing (Continued), fonttitle=\bfseries, colback=white]
\texttt{\textbf{"turn 2"}: \{ \\
\ \ \ \ \textbf{"caption"}: "Turn the pen into a blue one", \\
\ \ \ \ \textbf{"objects"}: [ \\
\ \ \ \ \ \ \ ["a blue pen", [97, 235, 162, 222], 1], \\
\ \ \ \ \ \ \ ["a spatula", [217, 55, 198, 232], 2]] \\
\ \ \ \ \textbf{"background"}: "empty background", \\
\ \ \ \ \textbf{"negative"}: "None" \\
\ \ \}, \\
\ \ \ \textbf{"turn 3"}: \{ \\
\ \ \ \ \textbf{"caption"}: "I don't want this anymore", \\
\ \ \ \ \textbf{"objects"}: [ \\
\ \ \ \ \ \ \ ["a spatula", [157, 140, 198, 232], 2]] \\
\ \ \ \ \textbf{"background"}: "empty background", \\
\ \ \ \ \textbf{"negative"}: "a blue pen" \\
\ \ \}, \\
\ \ \textbf{"turn 4"}: \{ \\
\ \ \ \ \textbf{"caption"}: "I want four of the remaining object.", \\
\ \ \ \ \textbf{"objects"}: [ \\
\ \ \ \ \ \ \ ["a spatula", [4, 20, 198, 232], 2], \\
\ \ \ \ \ \ \ ["a spatula", [219, 20, 198, 232], 2], \\
\ \ \ \ \ \ \ ["a spatula", [85, 260, 198, 232], 2], \\
\ \ \ \ \ \ \ ["a spatula", [310, 260, 198, 232], 2]] \\
\ \ \ \ \textbf{"background"}: "empty background", \\
\ \ \ \ \textbf{"negative"}: "None" \\
\ \ \} \\
\}}
\end{tcolorbox}

\vspace{5em}
\section{More Visualization Results}

\begin{figure*}[!t]
  \centering
  \includegraphics[width=\textwidth]{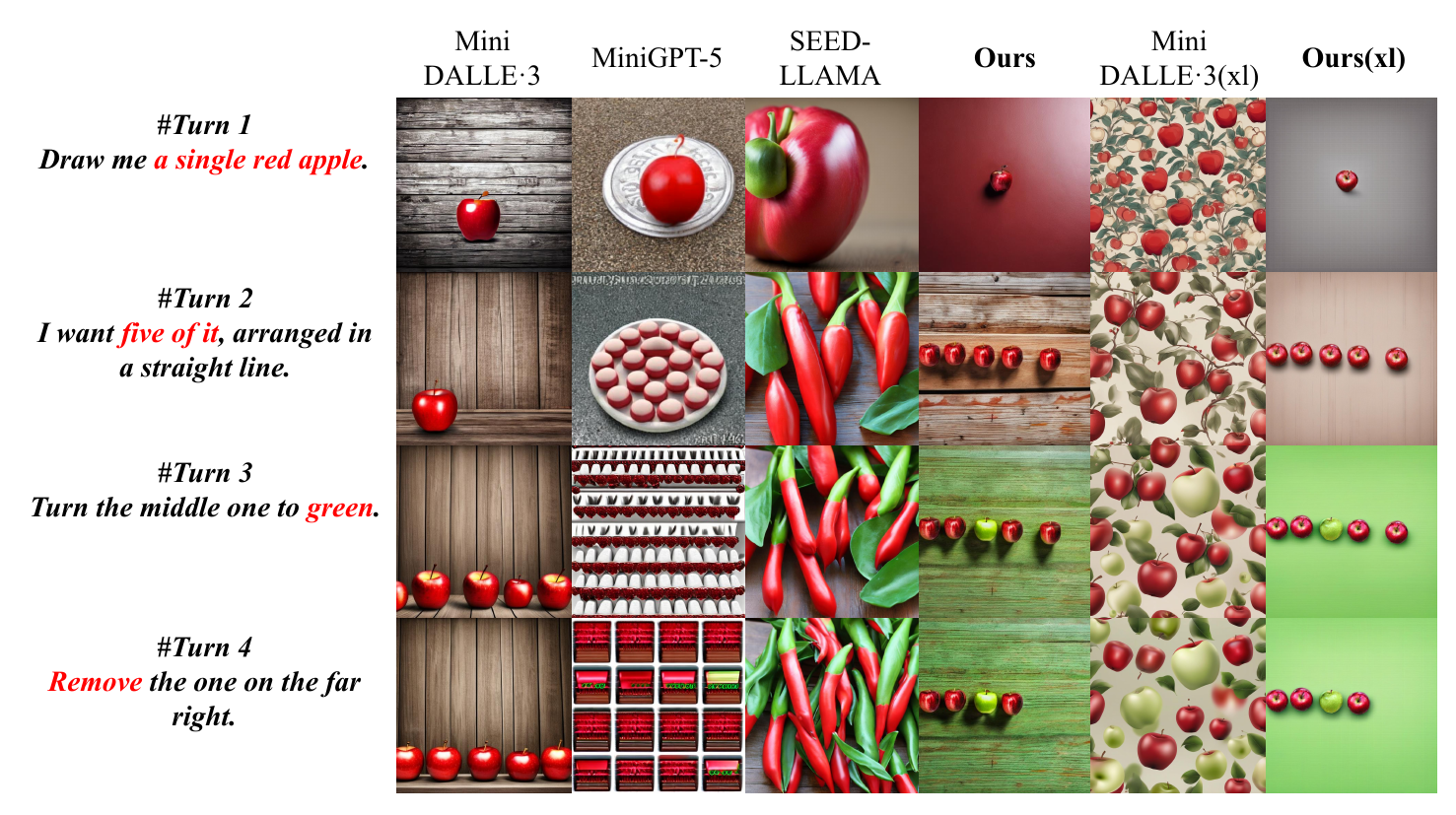}
  \caption{Visualization results of multi-turn editing.}
  \label{fig:edit 2}
\end{figure*}

\begin{figure*}[!t]
  \centering
  \includegraphics[width=\textwidth]{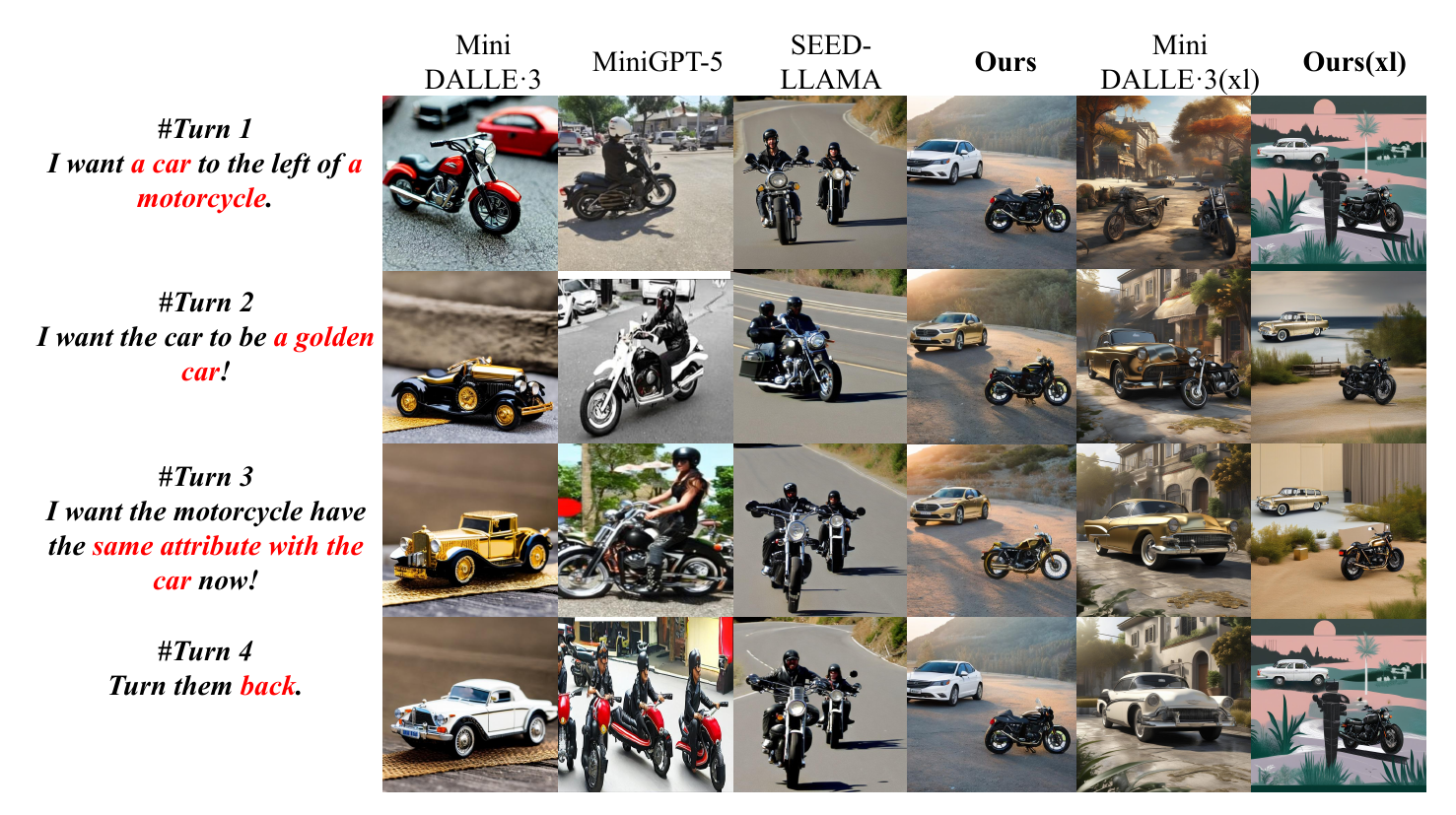}
  \caption{Visualization results of multi-turn editing.}
  \label{fig:edit 3}
\end{figure*}

\begin{figure*}[!t]
  \centering
  \includegraphics[width=\textwidth]{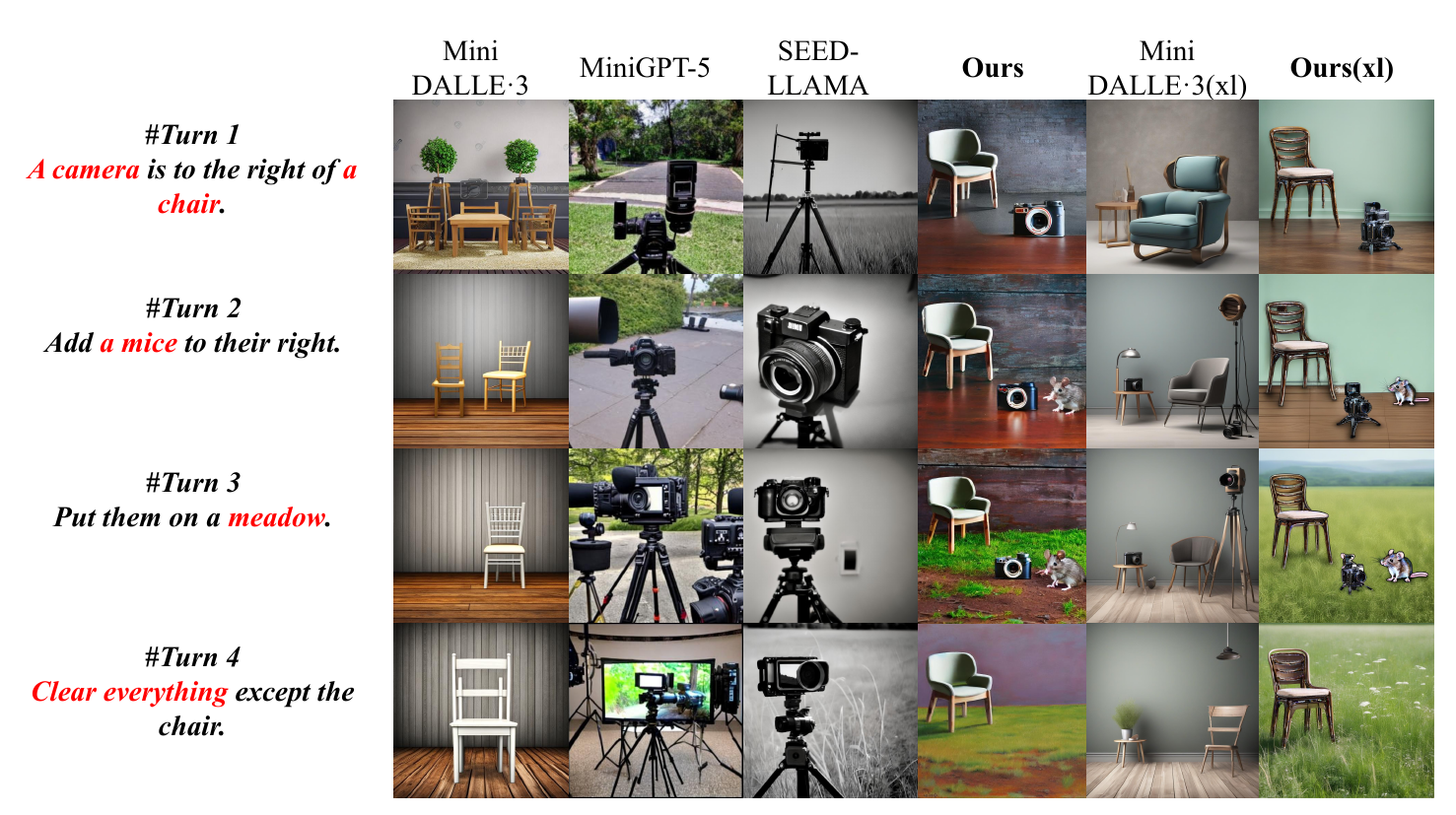}
  \caption{Visualization results of multi-turn editing.}
  \label{fig:edit 4}
\end{figure*}

\begin{figure*}[!t]
  \centering
  \includegraphics[width=\textwidth]{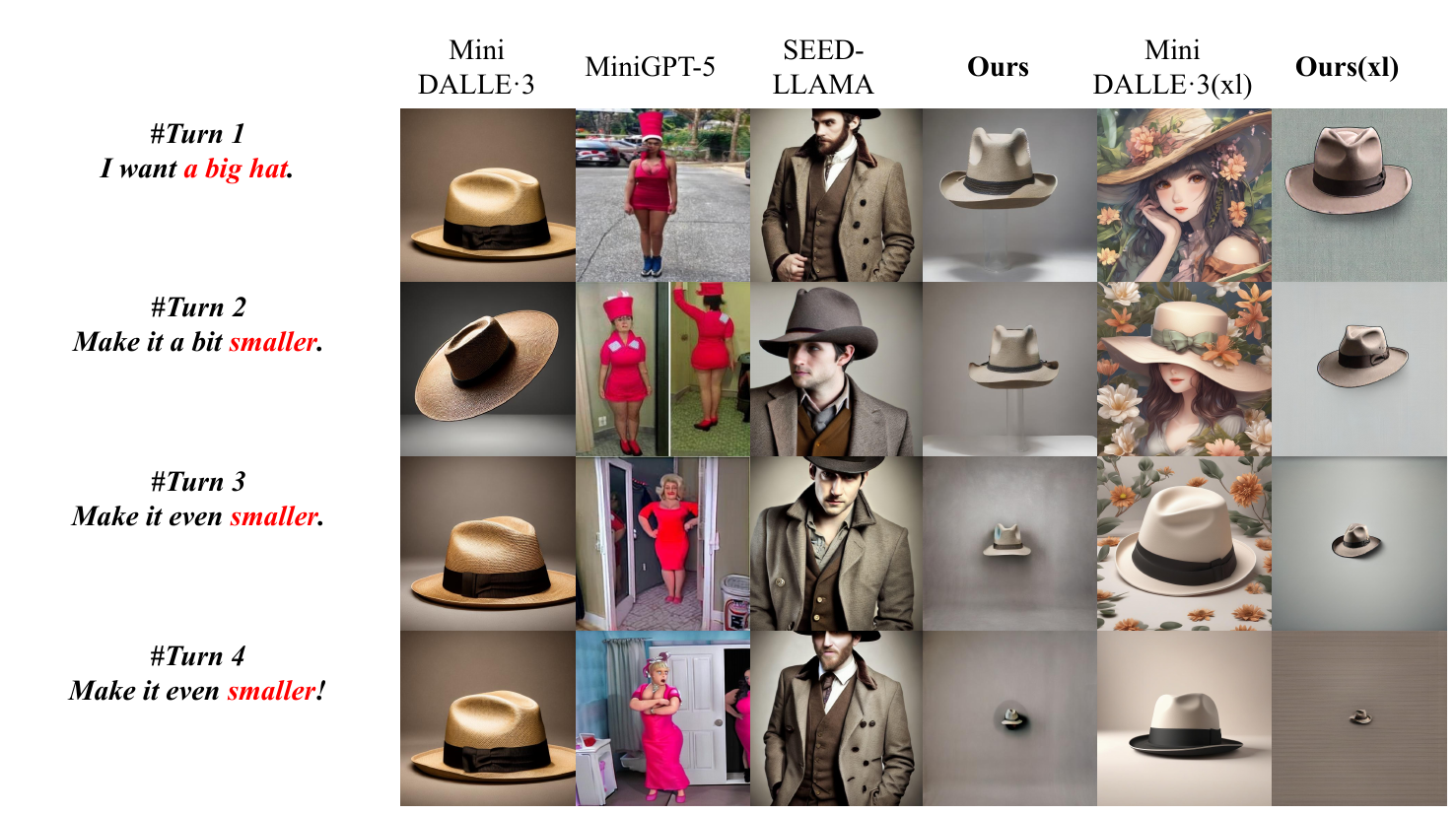}
  \caption{Visualization results of multi-turn editing.}
  \label{fig:edit 5}
\end{figure*}

\begin{figure*}[!t]
  \centering
  \includegraphics[width=\textwidth]{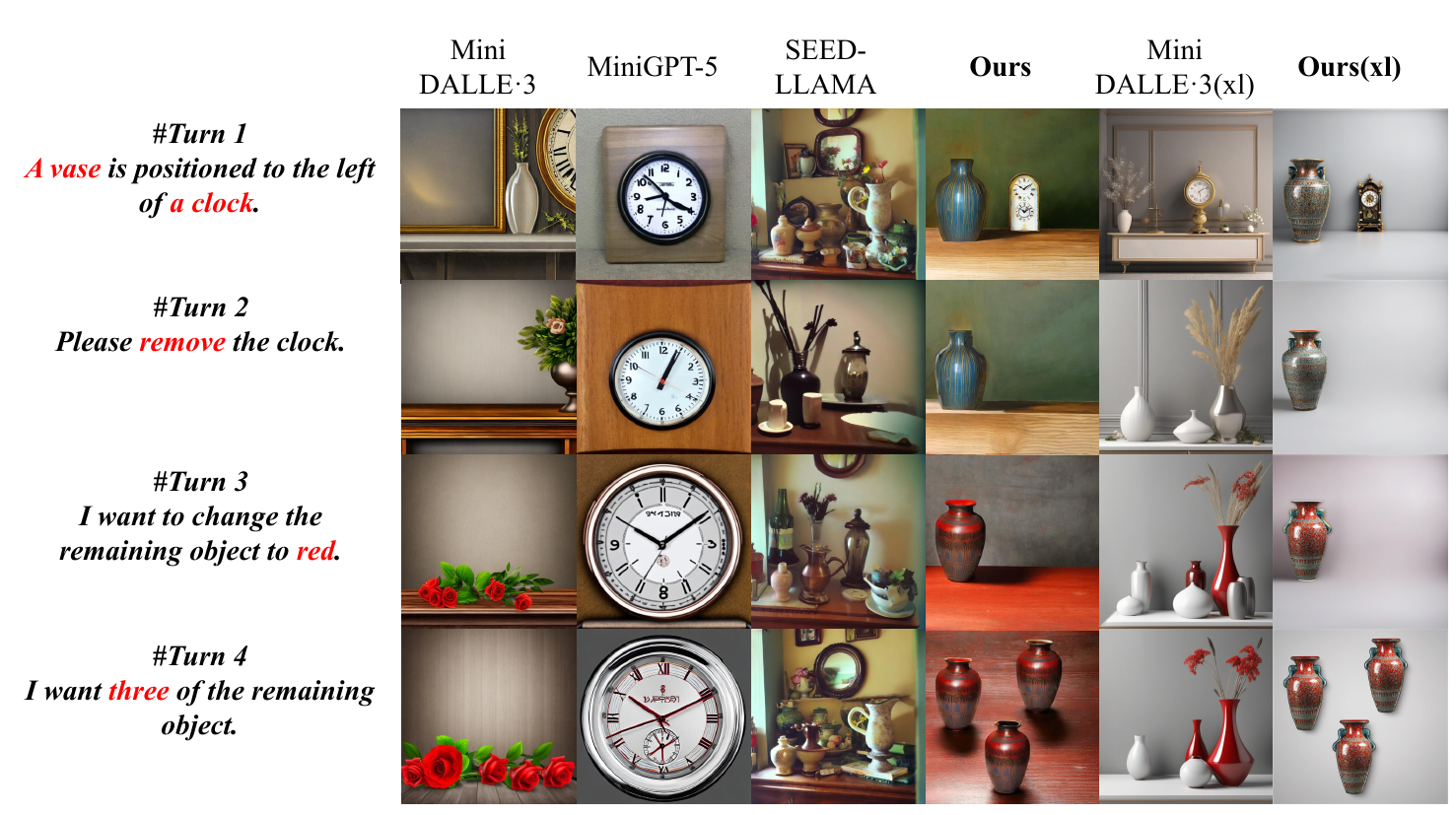}
  \caption{Visualization results of multi-turn editing.}
  \label{fig:edit 6}
\end{figure*}

\begin{figure*}[!t]
  \centering
  \includegraphics[width=\textwidth]{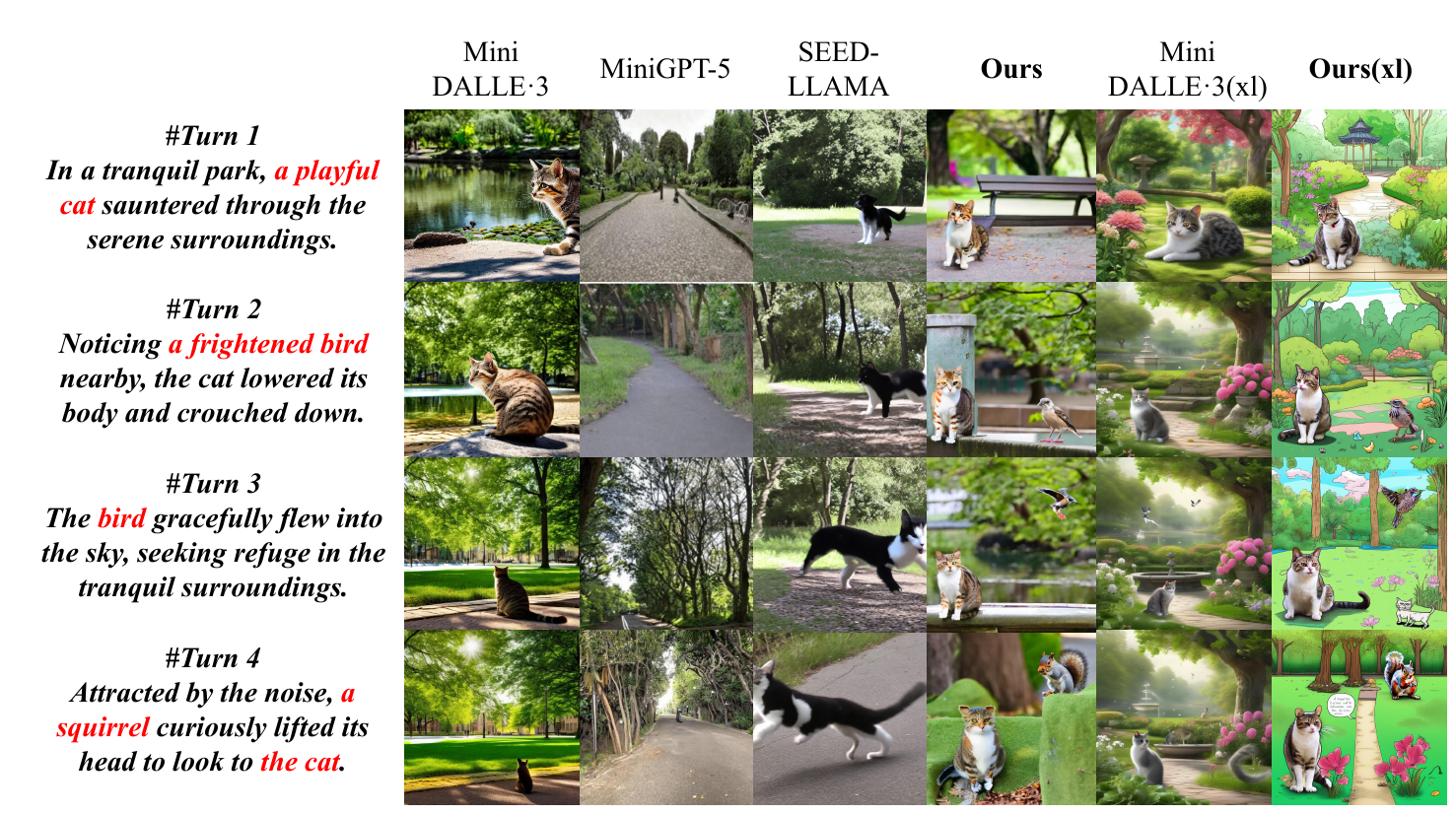}
  \caption{Visualization results of story generation.}
  \label{fig:story 7}
\end{figure*}

\begin{figure*}[!t]
  \centering
  \includegraphics[width=\textwidth]{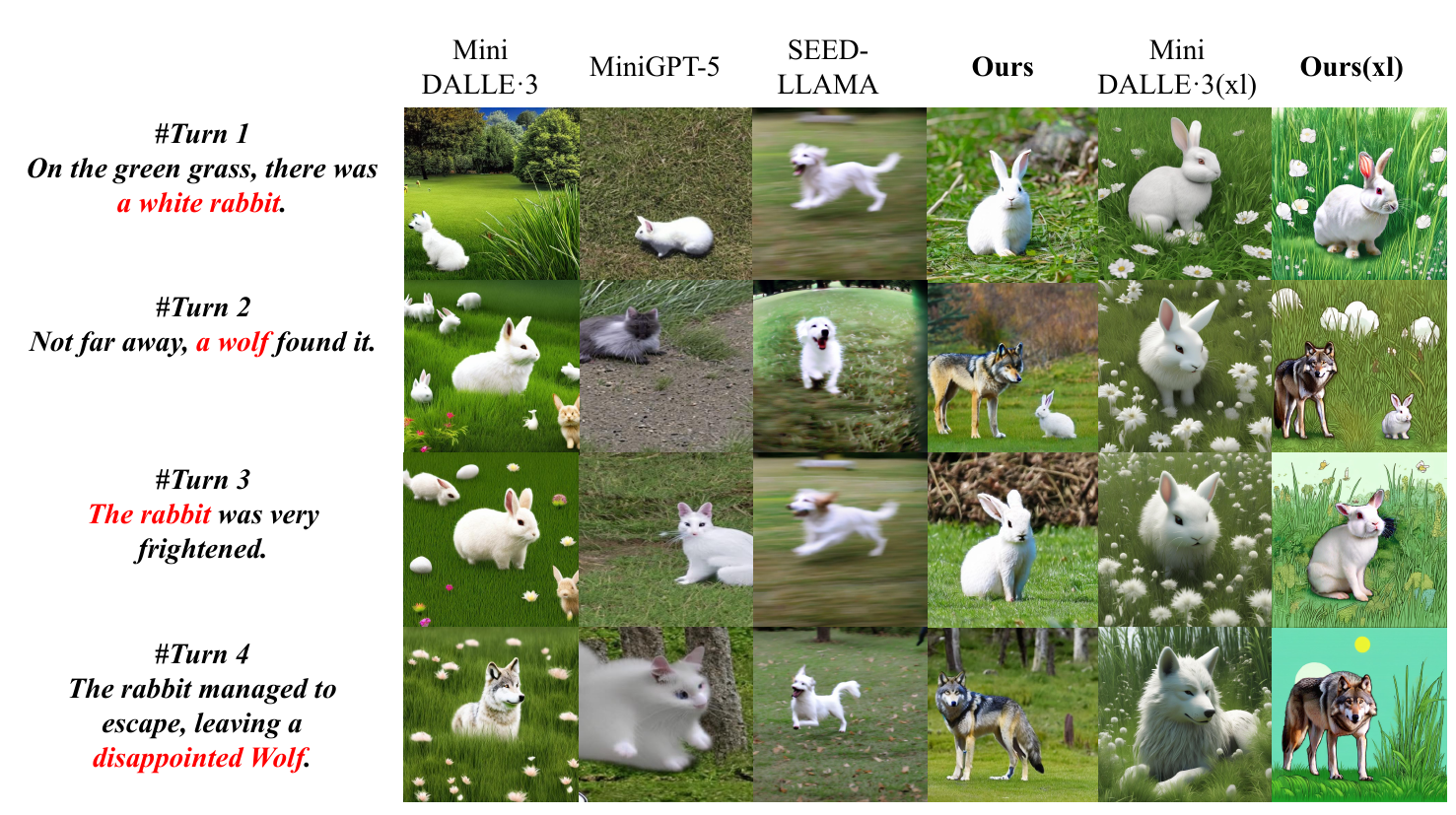}
  \caption{Visualization results of story generation.}
  \label{fig:story 8}
\end{figure*}

\begin{figure*}[!t]
  \centering
  \includegraphics[width=\textwidth]{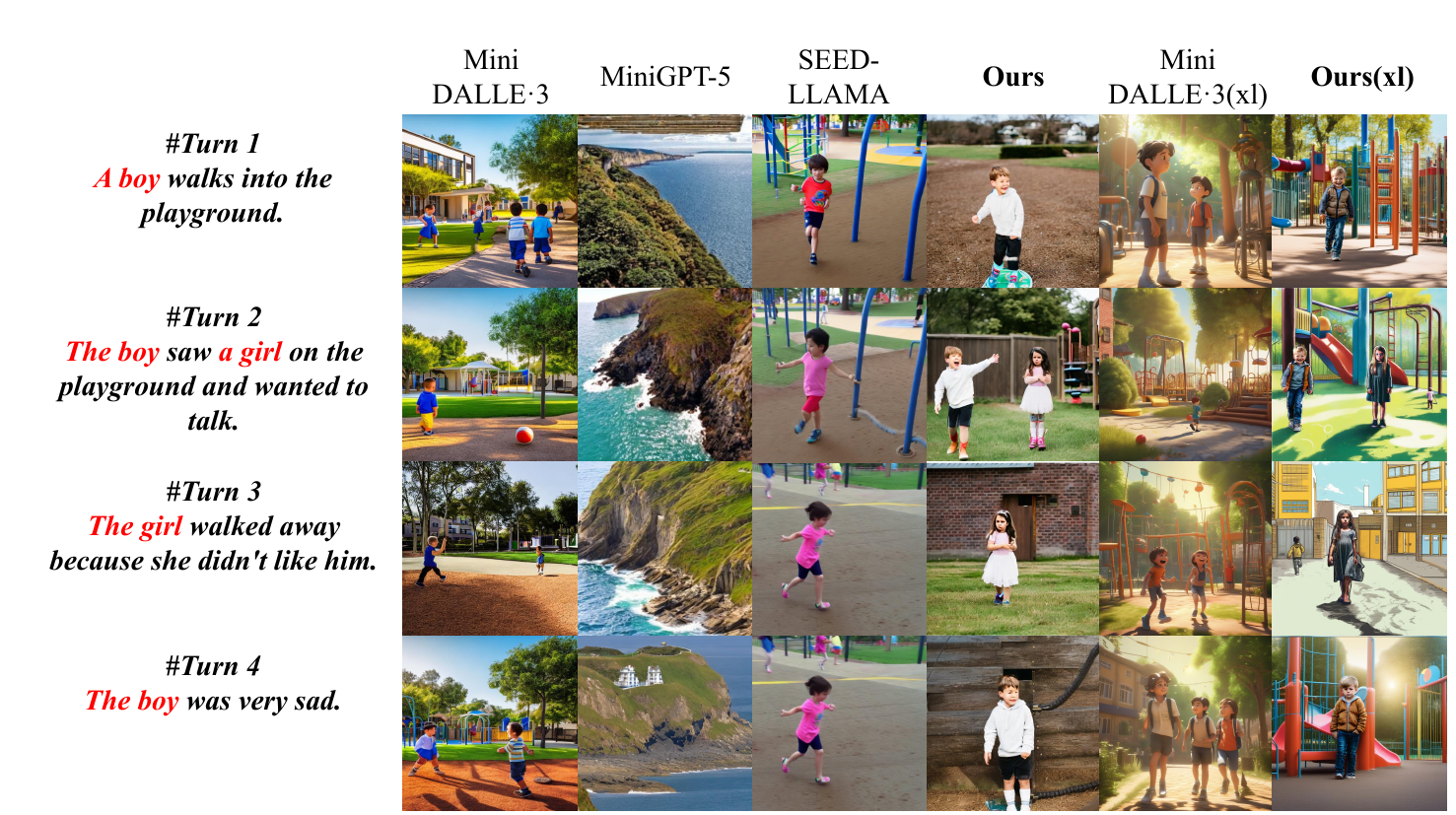}
  \caption{Visualization results of story generation.}
  \label{fig:story 9}
\end{figure*}

\begin{figure*}[!t]
  \centering
  \includegraphics[width=\textwidth]{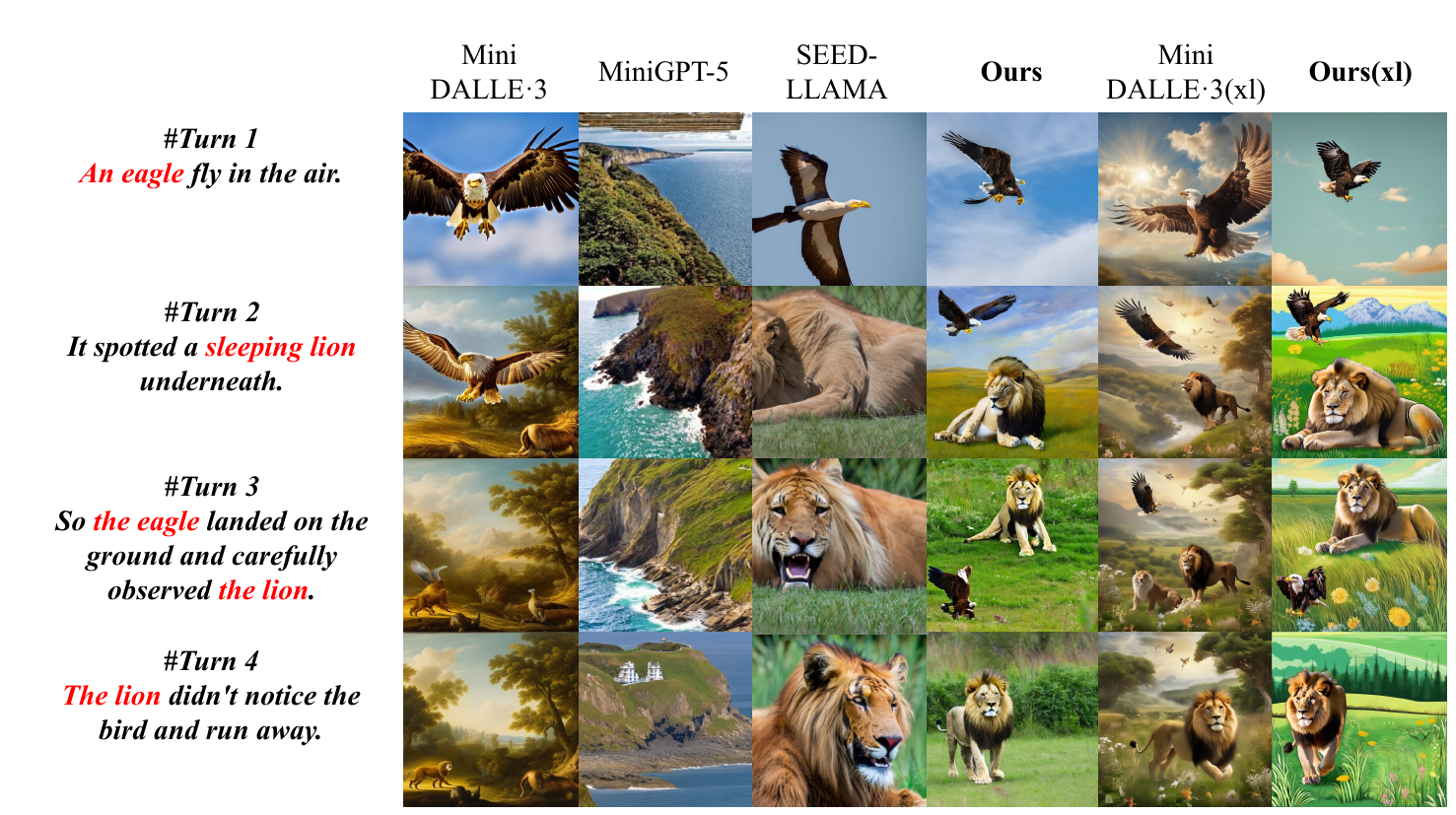}
  \caption{Visualization results of story generation.}
  \label{fig:story 10}
\end{figure*}

\begin{figure*}[!t]
  \centering
  \includegraphics[width=\textwidth]{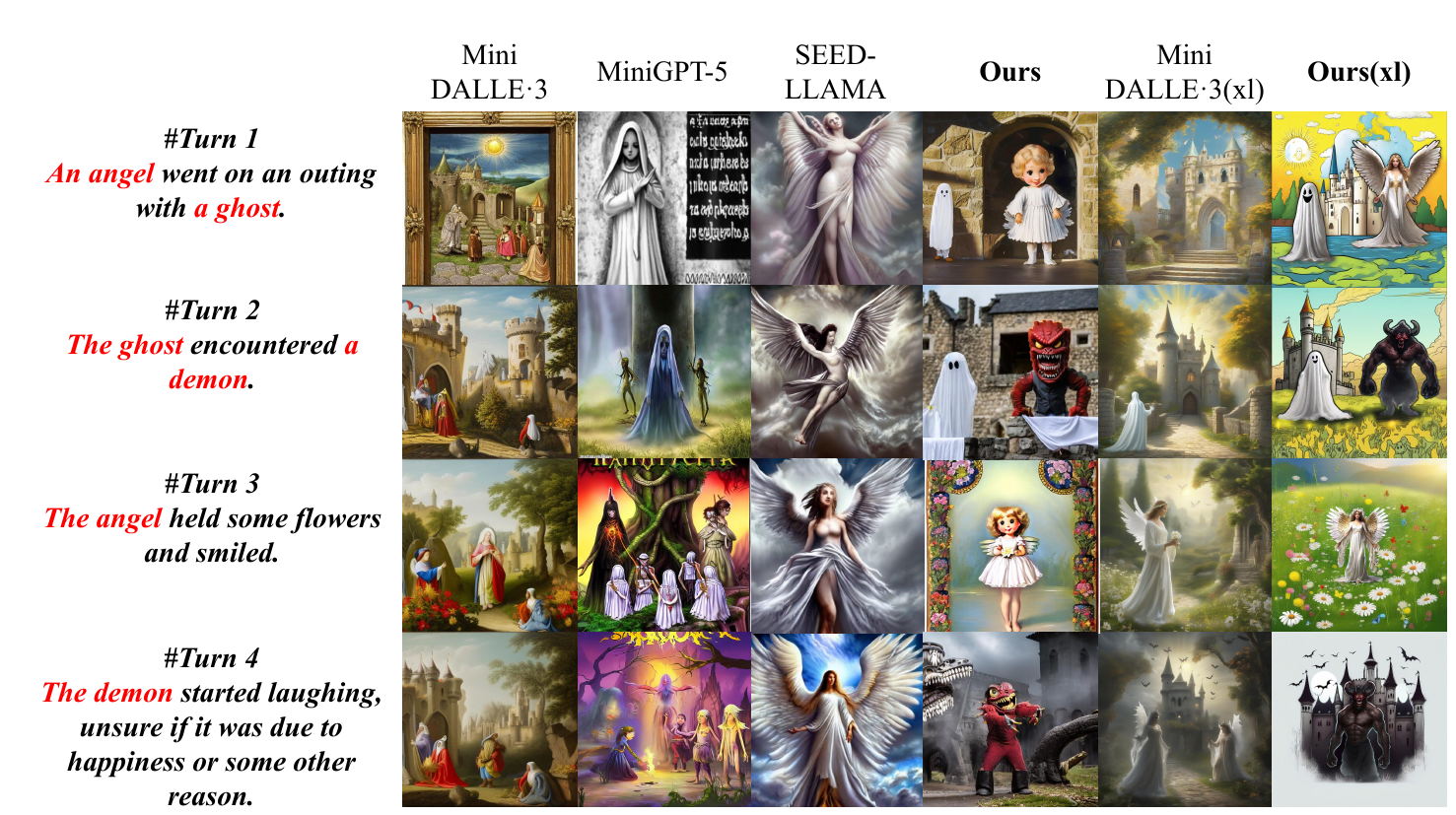}
  \caption{Visualization results of mstory generation.}
  \label{fig:story 11}
\end{figure*}

\begin{figure*}[!t]
  \centering
  \includegraphics[width=0.7\textwidth]{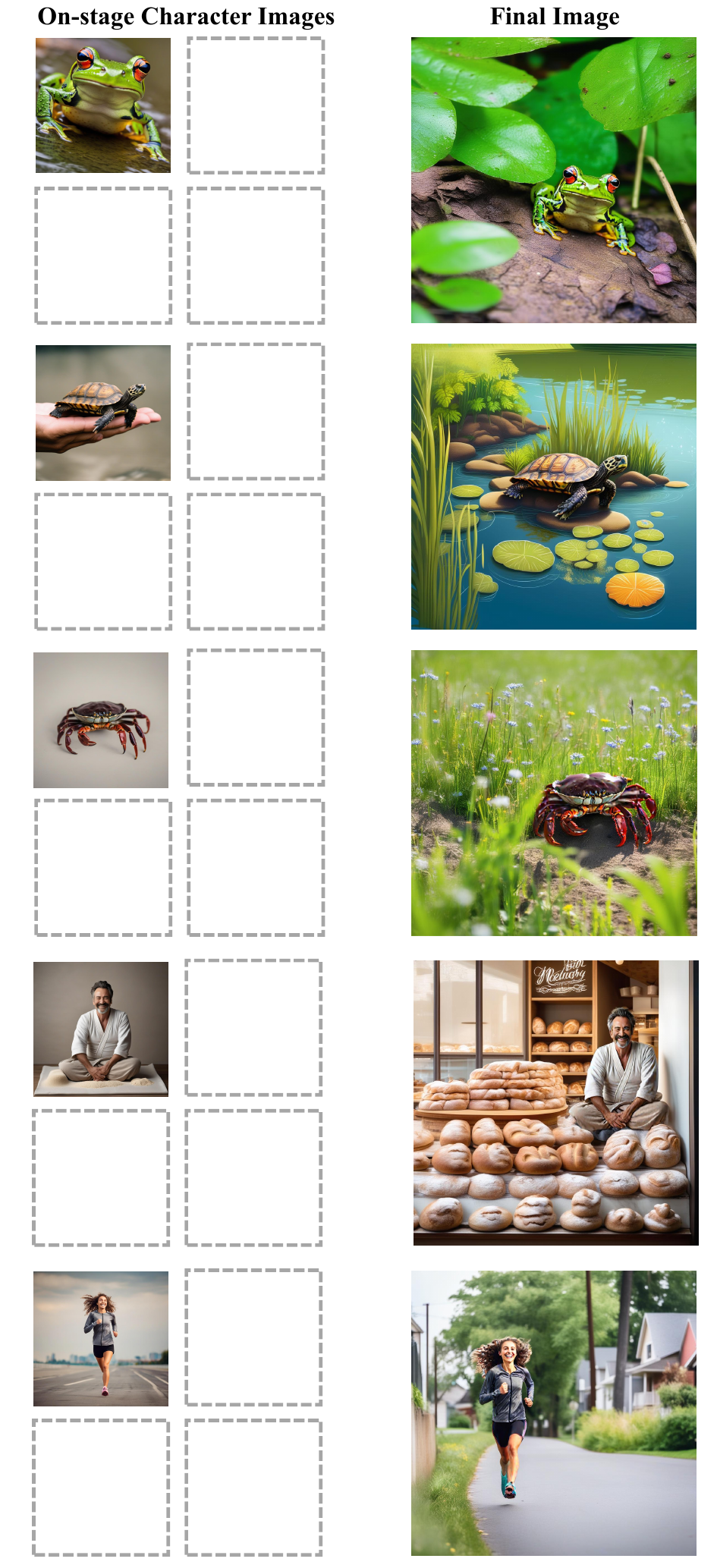}
  \caption{Comparison between on-stage character images and the final image.}
  \label{fig:big 1}
\end{figure*}

\begin{figure*}[!t]
  \centering
  \includegraphics[width=0.7\textwidth]{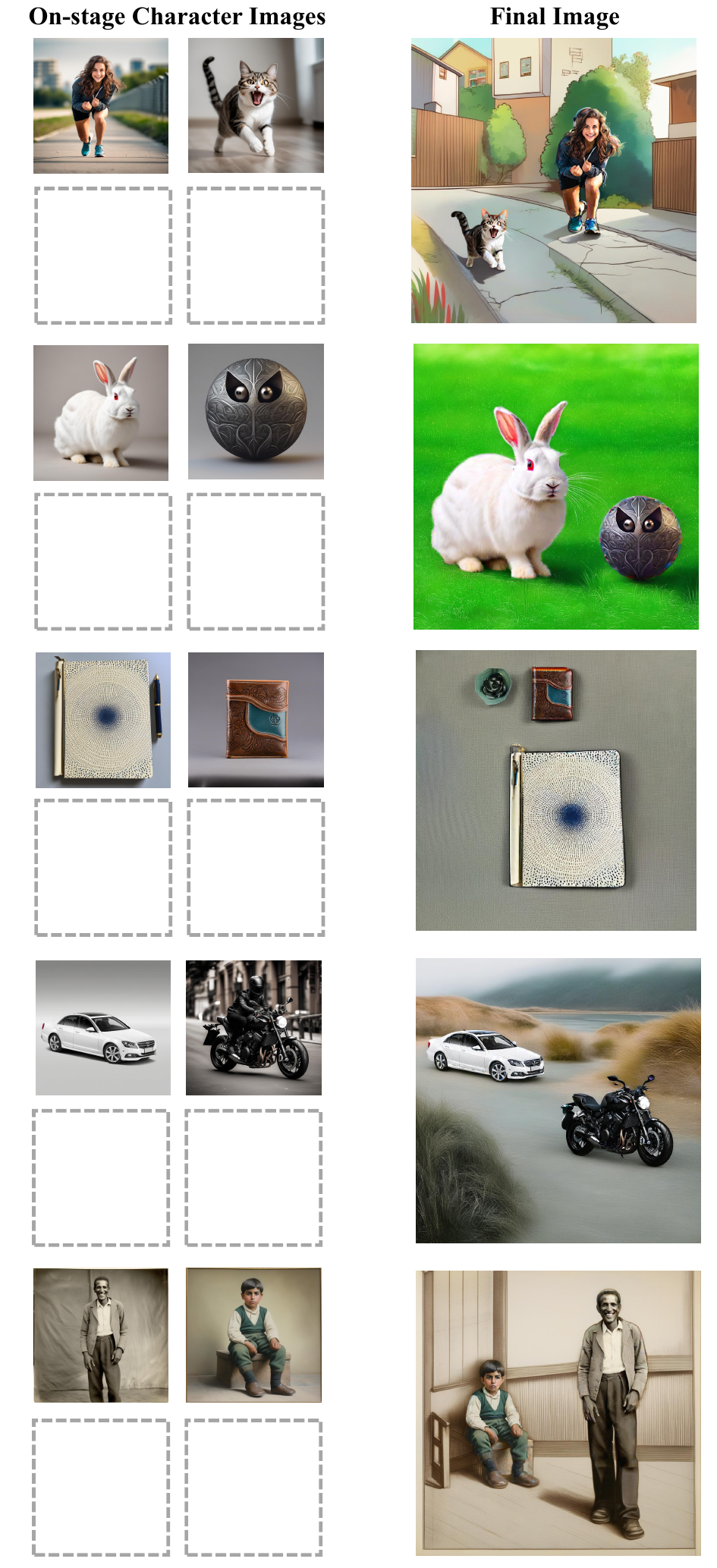}
  \caption{Comparison between on-stage character images and the final image.}
  \label{fig:big 2}
\end{figure*}

\begin{figure*}[!t]
  \centering
  \includegraphics[width=0.7\textwidth]{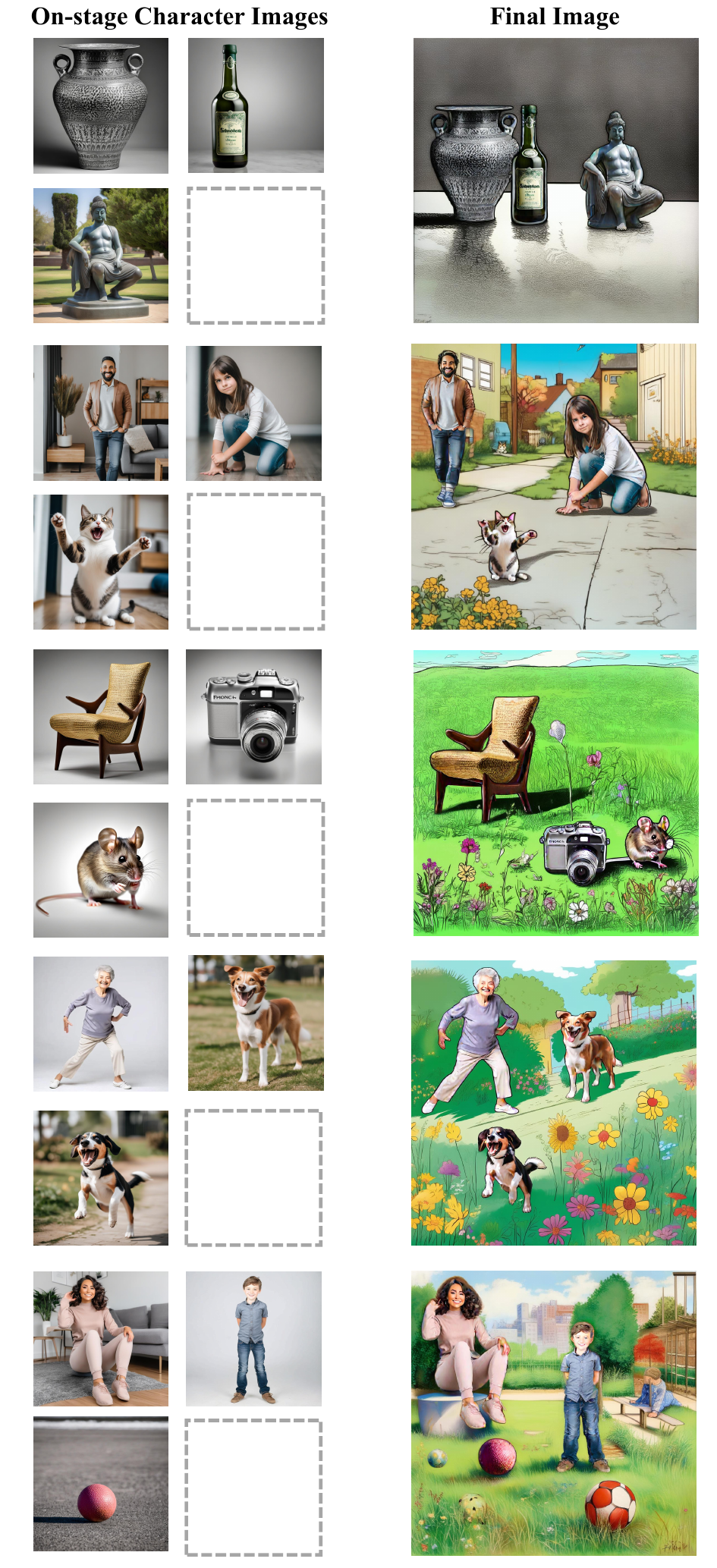}
  \caption{Comparison between on-stage character images and the final image.}
  \label{fig:big 3}
\end{figure*}

\begin{figure*}[!t]
  \centering
  \includegraphics[width=0.7\textwidth]{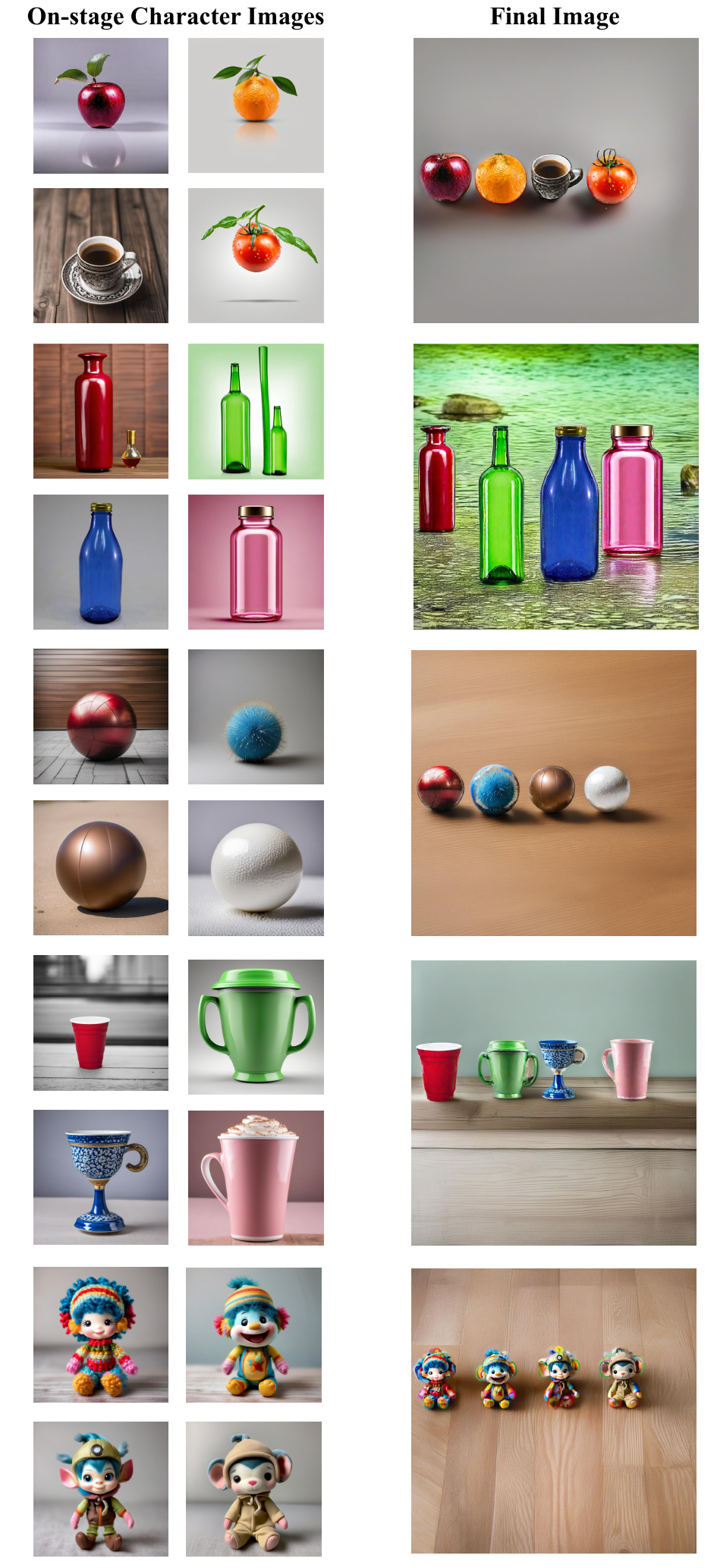}
  \caption{Comparison between on-stage character images and the final image.}
  \label{fig:big 4}
\end{figure*}

Figure~\ref{fig:edit 2} to Figure~\ref{fig:edit 6} showcase results of multi-turn editing. Figure~\ref{fig:story 7} to Figure~\ref{fig:story 11} showcase results of story generation. Figure~\ref{fig:big 1} to Figure~\ref{fig:big 4} showcases high-quality final images and the corresponding on-stage character images generated by TheaterGen based on SDXL. This demonstrates that TheaterGen is capable of preserving character features to the maximum extent while achieving a seamless integration of foreground and background.


\clearpage  
\bibliographystyle{splncs04}
\bibliography{main}

\end{document}